\documentclass[preprint,10pt,3p, times]{elsarticle}
\usepackage{amsmath,amsfonts}
\usepackage{algorithmic}
\usepackage{algorithm}
\usepackage{array}
\usepackage[caption=false,font=normalsize,labelfont=sf,textfont=sf]{subfig}
\usepackage{textcomp}
\usepackage{stfloats}
\usepackage{url}
\usepackage{verbatim}
\usepackage{graphicx}
\usepackage{cite}
\usepackage{multicol,multirow}
\usepackage{booktabs}
\usepackage{color}
\usepackage{hyperref}
\usepackage{xcolor} 
\usepackage[ss]{cmll} 

\usepackage{numprint}
\npdecimalsign{.}
\newcommand{\eer}[1]{\nprounddigits{2}\numprint{#1}}
\newcommand{\dcf}[1]{\nprounddigits{3}\numprint{#1}}

\newcommand{\stdev}[1]{{\scriptsize\color{gray}$\pm$#1}}
\newcommand{\nameproject}{\textit{LeBenchmark }}
\newcommand{\nameprojectnew}{\textit{LeBenchmark~2.0 }}

\newcommand{\oL}{1K-\textit{large} }
\newcommand{\tL}{3K-\textit{large} }

\definecolor{rebuttal}{HTML}{000000}
\definecolor{rebuttal2}{HTML}{000000}

\journal{Computer Speech \& Language}

\begin{document}

\begin{frontmatter}

\title{LeBenchmark 2.0: a Standardized, Replicable and Enhanced Framework for Self-supervised Representations of French Speech}

\author[samsung,cambridge]{Titouan Parcollet}
\author[avignon]{Ha~Nguyen}
\author[grenoble]{Solène~Evain}
\author[naver]{Marcely~Zanon~Boito}
\author[grenoble]{Adrien Pupier}
\author[avignon]{Salima~Mdhaffar}
\author[grenoble]{Hang~Le}
\author[grenoble]{Sina~Alisamir}
\author[avignon]{Natalia~Tomashenko}
\author[grenoble]{Marco~Dinarelli}
\author[samsung]{Shucong~Zhang}
\author[paris]{Alexandre~Allauzen}
\author[grenoble]{Maximin~Coavoux}
\author[avignon]{Yannick~Estève}
\author[avignon]{Mickael~Rouvier}
\author[grenoble]{Jerôme~Goulian}
\author[grenoble]{Benjamin~Lecouteux}
\author[grenoble]{François~Portet}
\author[grenoble]{Solange~Rossato}
\author[grenoble]{Fabien~Ringeval}
\author[grenoble]{Didier~Schwab}
\author[naver]{Laurent~Besacier}

\affiliation[samsung]{organization={Samsung AI Center Cambridge},
             addressline={50/60 Station Road},
             city={Cambridge},
             postcode={CB1 2JH},
             country={United Kingdom}}

\affiliation[cambridge]{organization={Department of Computer Science and Technology, University of Cambridge},
             addressline={15 JJ Thomson Av.},
             city={Cambridge},
             postcode={CB3 0FD},
             country={United Kingdom}}

\affiliation[avignon]{organization={Laboratoire Informatique d'Avignon, Avignon Université},
             addressline={339 Chem. des Meinajariès},
             city={Avignon},
             postcode={84000},
             country={France}}

\affiliation[grenoble]{organization={Univ. Grenoble Alpes, Inria, CNRS, Grenoble INP, LIG}, 
            postcode={38000},
            city={Grenoble}, 
            country={France}}

\affiliation[paris]{organization={ESPCI, CNRS LAMSADE, PSL Research University}, 
            country={France}}

\affiliation[naver]{organization={NAVER LABS Europe}, 
            country={France}}


\begin{abstract}

Self-supervised learning (SSL) is at the origin of unprecedented improvements in many different domains including computer vision and natural language processing. Speech processing drastically benefitted from SSL as most of the current domain-related tasks are now being approached with pre-trained models. This work introduces \nameprojectnew an open-source framework for assessing and building SSL-equipped French speech technologies. It includes documented, large-scale and heterogeneous corpora with up to 14,000 hours of heterogeneous speech, ten pre-trained SSL wav2vec 2.0 models containing from 26 million to one billion learnable parameters shared with the community, and an evaluation protocol made of six downstream tasks to complement existing benchmarks. \nameprojectnew also presents unique perspectives on pre-trained SSL models for speech with the investigation of frozen versus fine-tuned downstream models, task-agnostic versus task-specific pre-trained models as well as a discussion on the carbon footprint of large-scale model training. \textcolor{rebuttal}{Overall, the newly introduced models trained on 14,000 hours of French speech outperform multilingual and previous \nameproject SSL models across the benchmark but also required up to four times more energy for pre-training.}
\end{abstract}








                \end{frontmatter}

\section{Introduction}
\label{sec:introduction}

%

Throughout solving pretext tasks automatically extracted from massive unlabeled data, Self-Supervised Learning (SSL) powered deep learning systems deliver groundbreaking performance across a wide range of domains including audio, speech, and language processing \citep{mohamed2022self, liu2022audio, liu2021self}, computer vision \citep{jing2020self, reed2022self}, robotics \citep{sinha2022s4rl}, embedded devices and sensors \citep{wang2022self}, and medicine \citep{krishnan2022self, shurrab2022self}. In the specific context of speech processing, almost every sub-field has been largely impacted by newly available large pre-trained SSL models. Indeed, impressive improvements and state-of-the-art performance in competitive datasets have been reported for Automatic Speech Recognition (ASR) \citep{hsu2021hubert, baevski2020wav2vec, chen2021wavlm}, Automatic Emotion Recognition (AER) \citep{sarkar2020self, lebenchmark2, yang21c_interspeech}, Automatic Speaker Verification (ASV) \citep{chen2022large, zhang2021contrastive, yang21c_interspeech}, Automatic Speech Translation (AST) \citep{nguyen2020investigating,babu2021xls}, Spoken Language Understanding (SLU) \citep{yang21c_interspeech, liu2021tera, dinarelli22_interspeech, laperriere2022use}, Speech Enhancement (SE) \citep{huang2022investigating,sivaraman2022efficient, wang2020self}, Speech Separation (SS) \citep{huang2022investigating, tsai2022superb}, and many others. Despite most leaderboards being conceived around the English language, SSL has also been reported to be remarkably useful for under-resourced languages as demonstrated by A. Babu et al.  \citep{babu2021xls} and H. Nguyen et al.~\citep{nguyen2020investigating}, drastically increasing the accessibility to cutting-edge speech technologies across many languages.

Naturally, the flourishing field of SSL for speech calls for fair comparisons and standardized evaluation protocols properly assessing the added value of each newly introduced architecture. Following other early-adopter domains including Natural Language Processing (NLP) with, for instance, the GLUE \citep{wang2018glue} and SuperGLUE benchmarks \citep{wang2019superglue}, a first English-only evaluation suite appeared: SUPERB \citep{yang21c_interspeech}. In the latter, 13 different tasks based on well-known datasets have been bundled together to benchmark novel SSL models following a common fine-tuning evaluation protocol. Nonetheless, SUPERB does not standardize the pre-training process and hyperparameters, and models trained on hundreds of thousands of hours of speech appear in the same leaderboard as those learned with a few hundred hours or even different languages. SUPERB has been extended to generative tasks such as speech enhancement following the same standardized evaluation protocol in SUPERB-SG \citep{tsai2022superb} \textcolor{rebuttal}{as well as multiple languages in ML-SUPERB \citep{shi23g_interspeech} or specialised to Indian languages with Indicsuperb \citep{javed2023indicsuperb}}. \textcolor{rebuttal}{Others benchmarks targetting a specific task, such as SLUE for SLU \citep{shon2022slue}, have also been proposed.} \nameproject approached the issue of SSL benchmarking in French from a unified perspective by freezing the available pre-training data as well as the fine-tuning procedure \citep{lebenchmark1, lebenchmark2}. It also introduced a set of pre-trained SSL models available to the community including the largest and best-performing French SSL systems. Aside from these two attempts, most SSL models currently are being compared in an arbitrary and heterogeneous fashion across different pre-training and fine-tuning datasets, evaluation protocols, and hyperparameters tuning. The standardization and available resources revolving around SSL evaluation remain scarce, and it is crucial to the community that further efforts are put in those directions. Indeed, the scientific value of a released model may only be validated if proven against a rigorous, fair and replicable evaluation protocol.

With the first version of \nameproject \citep{lebenchmark1, lebenchmark2}, and following the definition of D. Schlangen \citep{schlangen2021targeting}, we aimed at providing the necessary foundations for the investigation and comparison of SSL models towards French-based downstream tasks. \textcolor{rebuttal}{Indeed language-specific SSL, as shown in \citep{lebenchmark1, lebenchmark2} and the present work, clearly outperform multilingual models depending on the downstream conditions.} \nameprojectnew builds upon the latter accomplishment to provide a standardized, fully replicable, and extended framework for the assessment and development of SSL representations of French speech. In particular, we release a well-curated pre-training set containing up to fourteen thousand hours of heterogeneous French speech (Section \ref{sec:data}), three novel pre-trained SSL models ranging from thirty million to one billion parameters (Section \ref{sec:models}), as well as two new evaluation tasks for ASV and syntactic analysis of spoken French (Section \ref{sec:benchmark}). \nameprojectnew also widens the discussions on the topics of the energy footprint of SSL models (Section \ref{sec:footprint}), the difference between language-specific and language-agnostic pre-training (Section \ref{sec:benchmark}). 
In short, \nameprojectnew is a collective attempt at unifying the community of SSL for the French language around common models, datasets and evaluation protocols.

\section{Evaluating SSL models with LeBenchmark 2.0: background and motivations}
\label{sec:background}

SSL for audio and speech is the process of defining and solving unsupervised proxy tasks, also referred to as pretext tasks or workers, motivated by the nature of the input signal itself. Proxy tasks define both an objective and a transformation applied to the training samples to extract the training targets. In practice, SSL models are first pre-trained following the latter strategy before turning into frozen or fine-tuned feature extractors for common supervised learning tasks. A major benefit of SSL approaches is the ability to leverage the ever-growing mass of available unlabeled data to drastically increase the performance observed on much more expensive and complex to obtain human-labeled tasks. In the context of audio and speech, SSL-based systems occupy the top ranks of most leaderboards and are widely adopted by the community with up to a tenth of recent proceedings from top-tier speech conferences (i.e. year 2023) containing at least one reference to SSL. 

SSL strategies for audio and speech may be divided into four different families: generative, predictive, contrastive, and multi-task. Generative methods aim at reconstructing the input audio signal after various corruptions ranging from masking to additive noises. For instance, Mockingjay \citep{liu2020mockingjay}, Tera \citep{liu2021tera}, DecoAR 2.0 \citep{ling2020decoar}, Speech-XLNet \citep{song2020speech}, MPC, pMPC \citep{yue2021phonetically} and data2vec \citep{baevski2022efficient} optimize their parameters towards the reconstruction or reordering of masked/shuffled input frames while Autoregressive Predictive Coding (APC) \citep{yang2022autoregressive} reconstructs the input signal. Predictive systems, including WavLM \citep{chen2021wavlm}, HuBERT \citep{hsu2021hubert}, or BEST-RQ \citep{chiu2022self} aim at predicting unsupervised discrete labels (e.g. clustering) obtained from the input samples. Contrastive approaches such as wav2vec 2.0 \citep{baevski2020wav2vec} or Contrastive Predicting Coding (CPC) \citep{oord2018representation}, on the other hand, optimize their latent representation to facilitate the distinction between positive and negative candidates originating from the given signal. Finally, multi-task SSL proposes to combine different objectives or modalities to build a rich feature extractor. For example, PASE+ \citep{ravanelli2020multi} merges up to ten different workers ranging from signal reconstruction to contrastive learning during the pre-training process. 

Such a rich landscape of models may be seen both as a curse and a blessing by the scientific community. It offers a wide range of possibilities and research directions but also suffers from a strong lack of evaluation standards. In fact, even the simple task of identifying the best-performing paradigm for a specific downstream task remains impossible with the current state of the art in SSL for audio and speech. Indeed, the construction and evaluation of SSL models may vary along numerous axes, hence drastically hindering the ease of comparison between novel solutions. \nameprojectnew specifically aims at standardizing those axes to speed up, facilitate, and democratize research around SSL pre-training in French.

More precisely, the life cycle of any SSL model is comprised of three major events: pre-training data gathering, training, and downstream evaluation. Ideally, the two latter steps should be merged, enabling the evaluation and comparison of SSL models at pre-training, hence alleviating a time-consuming downstream assessment. In practice, however, this idea appears as a major scientific and technical challenge as the literature relies entirely on the above-described three-step process. Unfortunately, each step may introduce important variations leading to heterogeneous and unreplicable evaluation protocols. For instance, PASE+ was trained on 100 hours of speech while HuBERT processed 60,000 hours, making it easy to define the best-performing model but practically impossible to distinguish the best pre-training strategy. Other variations include, but are not limited to: differences in pre-training languages and data type during step one~(e.g. spontaneous against read speech), compute resources at step two (e.g., a single Nvidia GTX 1080 Ti for Mockingjay against 128 Nvidia Tesla V100 for wav2vec 2.0) or the lack of standards during downstream fine-tuning at step three (e.g., fine-tuning against frozen feature extractors, pre-training dataset included or excluded from the downstream evaluation, or simply the list of downstream tasks to include). Ultimately, such requirements, and particularly the need for large compute resources limit the access to SSL pre-training research to a tiny subset of well-equipped institutions and companies, drastically limiting the exploration and emergence of novel paradigms. 

Aside from pre-training efficiency, the community naturally attempted to standardize the third step while developing and comparing their models. For instance, ASR evaluation using the Librispeech dataset can be found for MockingJay, wav2vec 2.0, HuBERT, or WavLM, while speaker recognition with VoxCeleb has been reported in PASE+ and MockingJay. Nonetheless, in most cases, the employed downstream architectures, evaluation protocols, or hyperparameters are entirely different, making it impossible to distinguish models that differ strongly in their pre-training process (e.g. PASE+ and HuBERT). This also prevents a strict comparison between close-performing models~(e.g. WavLM and HuBERT). 

The increasingly adopted SUPERB benchmark~\citep{yang21c_interspeech} defines a set of English downstream tasks to compare SSL models, hence facilitating step three. Despite a long list of 13 tasks, SUPERB suffers from a constrained fine-tuning procedure that forces all pre-trained SSL feature extractors to remain frozen and use a fixed decoder to solve the task of interest. Unfortunately, state-of-the-art SSL results and real-life use cases mostly, if not only, come with a joint fine-tuning of the SSL extractor and the decoder. S. Zaiem et al. \citep{zaiem2023benchmark} have also demonstrated that freezing all the downstream architectures and reducing them to a tiny subset could lead to misleading leaderboard rankings. Since the data preparation of step one is not standardized within SUPERB, it remains challenging to compare different SSL pre-training methodologies as the amount and quality of the data often vary between available SSL models. \nameproject is the first attempt at standardizing both steps one and three as well as providing replicable and competitive baselines for step two for further investigation from the community interested in the French language.

Finally, the current trend in large-scale SSL is to associate hundreds of languages \citep{babu2021xls} during pre-training without any regard to potential biases or degradation in performance induced by such a mixing.  However, it remains unclear if combining unrelated and potentially distant dialects may harm the performance observed compared to a model trained on a single and well-resourced language~(e.g. English). In particular, with \nameproject, we decided to benefit from the available unsupervised and heterogeneous French speech corpora available to train multiple language-specific SSL models \citep{lebenchmark1, lebenchmark2}, and we have demonstrated that such well-designed models usually outperform massively multilingual alternatives. Interestingly enough, and despite French being the fifth most spoken language, \nameproject is the only attempt at standardizing the data collection, pre-training, and evaluation phases of French SSL models. With \nameprojectnew we wish to further enhance our already adopted unified SSL framework for the French community, as both industry and academic institutions delivering state-of-the-art speech technologies are now building SSL-powered solutions. 

More precisely, \nameprojectnew extends \citep{lebenchmark1,lebenchmark2} in every aspect composing the framework and the three steps of the SSL life-cycle:
\begin{itemize}
    \item \textit{SSL data collection.} \citep{lebenchmark1} and \citep{lebenchmark2} offered carefully curated and documented corpora with 3,000 and 7,000 hours of French respectively. \nameprojectnew extends the openly available pretraining resources to 14,000 hours \textcolor{rebuttal}{thanks to a project-specific corpus: \textit{audiocite.net} (see sec.~\ref{sec:audiocite}), shared with }the same quality of documentation.
    \item \textit{SSL models pre-training.} \citep{lebenchmark1} and \citep{lebenchmark2} delivered up to seven pre-trained \textcolor{rebuttal}{wav2vec 2.0 SSL models} to the community based on the well-known Fairseq toolkit \citep{ott2019fairseq}. Following our newly introduced 14,000 hours of data, \nameprojectnew brings three more models, of which two are the largest ones available, to the community. Pre-training and model sharing are conducted with HuggingFace and SpeechBrain~\citep{ravanelli2021speechbrain}, two frameworks renowned for their open-science-minded approach. We also propose to extend the analysis and discussion on the energy footprint of large SSL models. \textcolor{rebuttal}{While we only offer pre-trained wav2vec 2.0 models due to the significant amount of compute resources and energy necessary to train them, we believe that the gathered datasets alongside the benchmarking tasks will emulate the community to extend the analysis to other SSL methods.}
    \item \textit{SSL Benchmarking.} \citep{lebenchmark1} and \citep{lebenchmark2} released four standardized tasks to evaluate and compare SSL models in French: ASR, AST, SLU, and AER. \nameprojectnew extends this evaluation protocol to six tasks with the introduction of automatic speaker verification and syntactic analysis. We also widened the comparison with the state-of-the-art, and language-specific against language-agnostic models.
\end{itemize}

\section{Gathering large collections of datasets}
\label{sec:data}
%

Up until recently, it was difficult to find publicly available large datasets of French speech (with the exception of EPAC). Recently, large multilingual corpora that include French have been made available, such as MLS~\citep{old_pratap_mls_2020}~(1,096\,h), or voxpopuli~\citep{wang2021voxpopuli}~(+4,500\,h). However, these are restricted to either read or well-prepared speech, failing to provide diversity in the speech samples, such as accented, spontaneous and/or affective speech. 
In this work, we gathered a large variety of speech corpora in French that cover different accents~(MLS, African Accented Speech, CaFE), acted emotions~(GEMEP, CaFE, Att-Hack), telephone dialogues~(PORTMEDIA), read~(MLS, African Accented French, MaSS) and spontaneous sentences~(CFPP2000, ESLO2, MPF, TCOF, NCCFr), broadcast speech~(EPAC) and professional speech~(Voxpopuli). 
Furthermore, to extend the amount of speech data used for pre-training by around 7k hours we also collected the audiocite.net dataset of non-professional read speech. \textcolor{rebuttal}{The details of its design can be found in section \ref{sec:audiocite}}.
Compared to MLS and Voxpopuli, our dataset is more diverse, carefully sourced and contains detailed metadata (speech type, and speaker gender).  
Moreover, it has a more realistic representation of speech turns in real life, compared to MLS and VoxPopuli. Each dataset is documented and can be accessed at least for research purposes.\footnote{Some of them being released by ELRA, they are available for a small fee.}  This section summarizes the datasets collected and how they were organized for the pre-training step and gives a short overview of the new \emph{audiocite.net} dataset.

\subsection{Overview of the Datasets Used for Pre-training}\label{subsec:overviewpretrainingdatasets}
Table~\ref{tab:medium_and_small_datasets} summarizes the statistics of the complete list of datasets considered for the study. The datasets have been organized in five main groups.\\ 

\begin{table*}[!htb]
\caption{Statistics for the speech corpora used to train SSL models according to gender information (male / female / unknown). 
The small dataset is from MLS only. Every dataset is composed of the previous one + additional data; 
duration: hour(s):minute(s).}
\centering
\scalebox{0.85}{
\scriptsize
\begin{tabular}{lcccccc}\toprule

\textbf{Corpus}   
& \textbf{License}
& \textbf{\# Utterances}               
& \textbf{Duration}                    
& \textbf{\# Speakers} 
& \textbf{Mean Utt. Duration} 
& \textbf{Speech type}\\

\midrule
\multicolumn{7}{c}{\textbf{Small dataset -- 1K}} \\ \hline

MLS French~\citep{old_pratap_mls_2020}  
& \textcolor{rebuttal}{CC BY 4.0}
& \begin{tabular}[c]{@{}c@{}}\textbf{263,055} \\ 124,590 / 138,465 / --\end{tabular}                   
& \begin{tabular}[c]{@{}c@{}}\textbf{1,096:43} \\ 520:13 / 576:29 / --\end{tabular}                 
& \begin{tabular}[c]{@{}c@{}}\textbf{178} \\ 80 / 98 / --\end{tabular}                
& \begin{tabular}[c]{@{}c@{}}\textbf{15\,s} \\ 15\,s / 15\,s / --\end{tabular}       
&Read\\\bottomrule\\

\multicolumn{7}{c}{\textbf{Medium-clean dataset -- 2.7K}} \\ \hline

EPAC**~\citep{old_esteve_epac_2010}   
& \textcolor{rebuttal}{ELRA NC}
& \begin{tabular}[c]{@{}c@{}}\textbf{623,250} \\ 465,859 / 157,391 / --\end{tabular}                  
& \begin{tabular}[c]{@{}c@{}}\textbf{1,626:02} \\ 1,240:10 / 385:52 / --\end{tabular}                
& \begin{tabular}[c]{@{}c@{}}\textbf{Unk} \\ -- / -- / --\end{tabular}           
& \begin{tabular}[c]{@{}c@{}}\textbf{9\,s} \\ -- / -- / --\end{tabular}                       
& \begin{tabular}[c]{@{}c@{}}Radio \\ Broadcasts\end{tabular}          \\\hline

\textbf{2.7k dataset total}  
&
& \begin{tabular}[c]{@{}c@{}}\textbf{886,305} \\ 590,449 / 295,856 / -\end{tabular} 
& \begin{tabular}[c]{@{}c@{}}\textbf{2,722:45} \\ 1,760:23 / 962:21 / -\end{tabular} 
& -                                                             
& -      
& - \\\bottomrule\\

\multicolumn{7}{c}{\textbf{Medium dataset -- 3K}} \\ \hline

\begin{tabular}[l]{@{}l@{}}African Accented \\ French~\citep{noauthor_african_2003}\end{tabular} 
& \textcolor{rebuttal}{Apache 2.0}
& \begin{tabular}[c]{@{}c@{}}\textbf{16,402} \\ 373 / 102 / 15,927\end{tabular}  
& \begin{tabular}[c]{@{}c@{}}\textbf{18:56} \\ -- / -- / 18:56\end{tabular}                      
& \begin{tabular}[c]{@{}c@{}}\textbf{232} \\ 48 / 36 / 148\end{tabular}         
& \begin{tabular}[c]{@{}c@{}}\textbf{4\,s} \\ -- / -- / --\end{tabular}
&  Read\\\hline

Att-Hack~\citep{le_moine_att-hack_2020}    
& \textcolor{rebuttal}{CC BY-NC-ND}
& \begin{tabular}[c]{@{}c@{}}\textbf{36,339} \\ 16,564 / 19,775 /  --\end{tabular}       
& \begin{tabular}[c]{@{}c@{}}\textbf{27:02} \\ 12:07 / 14:54 / --\end{tabular}     
& \begin{tabular}[c]{@{}c@{}}\textbf{20} \\ 9 / 11 / -- \end{tabular}
& \begin{tabular}[c]{@{}c@{}}\textbf{2.7\,s}\\ 2.6\,s / 2.7\,s / --\end{tabular}    
& \begin{tabular}[c]{@{}c@{}}Acted \\ Emotional\end{tabular}          \\\hline

CaFE~\citep{gournay_canadian_2018} 
& \textcolor{rebuttal}{CC NC}
& \begin{tabular}[c]{@{}c@{}}\textbf{936} \\ 468 / 468 / --\end{tabular}            
& \begin{tabular}[c]{@{}c@{}}\textbf{1:09} \\ 0:32 / 0:36 / --\end{tabular}       
& \begin{tabular}[c]{@{}c@{}}\textbf{12} \\ 6 / 6 / --\end{tabular} 
& \begin{tabular}[c]{@{}c@{}}\textbf{4.4\,s} \\ 4.2\,s / 4.7\,s / --\end{tabular}     
& \begin{tabular}[c]{@{}c@{}}Acted \\ Emotional\end{tabular}          \\\hline

\begin{tabular}[c]{@{}c@{}}CFPP2000*\\ \citep{branca-rosoff_discours_2012}\end{tabular} 
& \textcolor{rebuttal}{CC BY-NC-SA}
& \begin{tabular}[c]{@{}c@{}}\textbf{9853} \\ 166 / 1,184 / 8,503\end{tabular}    
& \begin{tabular}[c]{@{}c@{}}\textbf{16:26}\\ 0:14 / 1:56 / 14:16\end{tabular}   
& \begin{tabular}[c]{@{}c@{}}\textbf{49} \\ 2 / 4 / 43\end{tabular}  
& \begin{tabular}[c]{@{}c@{}}\textbf{6\,s}\\ 5\,s / 5\,s / 6\,s\end{tabular}     
& Spontaneous\\\hline

ESLO2~\citep{old_eshkol-taravella_grand_2011} 
& \textcolor{rebuttal}{CC BY-NC-SA}
& \begin{tabular}[c]{@{}c@{}}\textbf{62,918} \\ 30,440 / 32,147 / 331\end{tabular}                          
& \begin{tabular}[c]{@{}c@{}}\textbf{34:12} \\ 17:06 / 16:57 / 0:09 \end{tabular}                          
& \begin{tabular}[c]{@{}c@{}}\textbf{190} \\ 68 / 120 / 2\end{tabular}                
& \begin{tabular}[c]{@{}c@{}}\textbf{1.9\,s} \\ 2\,s / 1.9\,s / 1.7\,s\end{tabular}                        
& Spontaneous           \\\hline

GEMEP~\citep{old_banziger_introducing_2012} 
& \textcolor{rebuttal}{User agreement}
& \begin{tabular}[c]{@{}c@{}}\textbf{1,236} \\ 616 / 620 / --\end{tabular}   
& \begin{tabular}[c]{@{}c@{}}\textbf{0:50} \\ 0:24 / 0:26 / --\end{tabular}      
& \begin{tabular}[c]{@{}c@{}} \textbf{10} \\ 5 / 5 / --\end{tabular} 
& \begin{tabular}[c]{@{}c@{}}\textbf{2.5\,s} \\ 2.4\,s / 2.5\,s / --\end{tabular}   
& \begin{tabular}[c]{@{}c@{}}Acted \\ Emotional\end{tabular} \\\hline

MPF~\citep{MPF_paper_2017}, \citep{MPF_ortolang}  
& \textcolor{rebuttal}{CC BY-NC-SA 4.0}
& \begin{tabular}[c]{@{}c@{}}\textbf{19,527} \\ 5,326 / 4,649 / 9,552\end{tabular}    
& \begin{tabular}[c]{@{}c@{}}\textbf{19:06} \\ 5:26 / 4:36 / 9:03\end{tabular}    
& \begin{tabular}[c]{@{}c@{}}\textbf{114}\\ 36 / 29 / 49\end{tabular}
& \begin{tabular}[c]{@{}c@{}}\textbf{3.5\,s} \\ 3.7\,s / 3.6\,s / 3.4\,s\end{tabular} 
& Spontaneous   \\\hline

\begin{tabular}[c]{@{}c@{}}PORTMEDIA\\ (French)~\citep{old_lefevre_robustesse_2012}\end{tabular} 
& \textcolor{rebuttal}{ELRA NC}
& \begin{tabular}[c]{@{}c@{}}\textbf{19,627} \\ 9,294 / 10,333 / --\end{tabular}  
& \begin{tabular}[c]{@{}c@{}}\textbf{38:59}\\ 19:08 / 19:50 / --\end{tabular}    
& \begin{tabular}[c]{@{}c@{}}\textbf{193} \\ 84 / 109 / --\end{tabular} 
& \begin{tabular}[c]{@{}c@{}}\textbf{7.1\,s} \\ 7.4\,s / 6.9\,s / --\end{tabular}   
& \begin{tabular}[c]{@{}c@{}}Acted telephone \\ dialogue \end{tabular} \\\hline

\begin{tabular}[c]{@{}c@{}}TCOF\\(Adults)~\citep{TCOF_ortolang}\end{tabular} 
& \textcolor{rebuttal}{CC BY-NC-SA}
& \begin{tabular}[c]{@{}c@{}}\textbf{58,722} \\ 10,377 / 14,763 / 33,582\end{tabular}  
& \begin{tabular}[c]{@{}c@{}}\textbf{53:59} \\ 9:33 / 12:39 / 31:46\end{tabular}  
& \begin{tabular}[c]{@{}c@{}}\textbf{749} \\ 119 / 162 / 468\end{tabular} 
& \begin{tabular}[c]{@{}c@{}}\textbf{3.3\,s} \\ 3.3\,s / 3.1\,s / 3.4\,s\end{tabular}  
& Spontaneous  \\\hline

\textbf{Medium dataset total}    
&
& \begin{tabular}[c]{@{}c@{}}\textbf{1,111,865} \\ 664,073 / 379,897 / 67,895\end{tabular} 
& \begin{tabular}[c]{@{}c@{}}\textbf{2,933:24} \\ 1,824:53 / 1,034:15 / 74:10\end{tabular} 
& -               
& -          
& -  \\\bottomrule\\

\multicolumn{7}{c}{\textbf{Large dataset -- 7K}} \\ \hline

MaSS~\citep{boito-etal-2020-mass} 
& \textcolor{rebuttal}{MIT}
& \begin{tabular}[c]{@{}c@{}}\textbf{8,219} \\ 8,219 / -- / -- \end{tabular}            
& \begin{tabular}[c]{@{}c@{}}\textbf{19:40} \\ 19:40 / -- / --\end{tabular} 
& \begin{tabular}[c]{@{}c@{}} \textbf{Unk} \\ -- / -- / --\end{tabular}   
& \begin{tabular}[c]{@{}c@{}}\textbf{8.6\,s} \\ 8.6\,s / -- / --\end{tabular}  
& Read    \\\hline

NCCFr~\citep{torreira_nijmegen_2010} 
& \textcolor{rebuttal}{User agreement}
& \begin{tabular}[c]{@{}c@{}}\textbf{29,421} \\ 14,570 / 13,922 / 929 \end{tabular}   
& \begin{tabular}[c]{@{}c@{}}\textbf{26:35} \\ 12:44 / 12:59 / 00:50\end{tabular}    
& \begin{tabular}[c]{@{}c@{}}\textbf{46}\\ 24 / 21 / 1\end{tabular}  
& \begin{tabular}[c]{@{}c@{}}\textbf{3\,s} \\ 3\,s / 3\,s / 3\,s \end{tabular} 
& Spontaneous  \\\hline

\begin{tabular}[l]{@{}l@{}}Voxpopuli~\citep{wang2021voxpopuli}\\\textit{Unlabeled}\end{tabular}
& \textcolor{rebuttal}{CC0}
& \begin{tabular}[c]{@{}c@{}}\textbf{568,338} \\ -- / -- / --\end{tabular}  
& \begin{tabular}[c]{@{}c@{}}\textbf{4,532:17}\\ -- / -- / 4,532:17\end{tabular}    
& \begin{tabular}[c]{@{}c@{}}\textbf{Unk} \\ -- / -- / --\end{tabular} 
& \begin{tabular}[c]{@{}c@{}}\textbf{29\,s} \\ -- / -- / --\end{tabular} 
& \begin{tabular}[c]{@{}c@{}}Professional speech \end{tabular} \\\hline
    
\begin{tabular}[l]{@{}l@{}}Voxpopuli~\citep{wang2021voxpopuli}\\\textit{transcribed}\end{tabular}
& \textcolor{rebuttal}{CC0}
& \begin{tabular}[c]{@{}c@{}}\textbf{76.281} \\ -- / -- / --\end{tabular}  
& \begin{tabular}[c]{@{}c@{}}\textbf{211:57}\\ -- / -- / 211:57\end{tabular}   
& \begin{tabular}[c]{@{}c@{}}\textbf{327} \\ -- / -- / --\end{tabular} 
& \begin{tabular}[c]{@{}c@{}}\textbf{10\,s} \\ -- / -- / --\end{tabular}  
& \begin{tabular}[c]{@{}c@{}}Professional speech \end{tabular} \\\midrule

\textbf{Large dataset total***}  
&
& \begin{tabular}[c]{@{}c@{}}\textbf{1,814,242} \\ 682,322 / 388,217 / 99,084\end{tabular} 
& \begin{tabular}[c]{@{}c@{}}\textbf{7,739:22} \\ 1,853:02 / 1,041:07 / 4,845:07\end{tabular} 
& -   
& -   
& -  \\\bottomrule\\

\multicolumn{7}{c}{\textbf{Extra Large dataset -- 14K}} \\ \hline
                                                    
\begin{tabular}[l]{@{}l@{}}Audiocite.net \\ (SLR139)~\citep{EVAINETAL_2022Audiocite}\end{tabular}
& \textcolor{rebuttal}{CC BY + ND/NC/SA}
& \begin{tabular}[c]{@{}c@{}}\textbf{817\,295} \\ 425\,033 / 159\,691 / 232\,571\end{tabular} 
& \begin{tabular}[c]{@{}c@{}}\textbf{6698:35}\\ 3477:24 / 1309:49 / 1911:21 \end{tabular}
& \begin{tabular}[c]{@{}c@{}}\textbf{130} \\ 35 / 32 / 63\end{tabular}  
& \begin{tabular}[c]{@{}c@{}}\textbf{29\,s} \\ 29\,s / 29\,s / 29\,s\end{tabular}                      
& \begin{tabular}[c]{@{}c@{}} Read \end{tabular} \\\midrule

\begin{tabular}[l]{@{}l@{}}Niger-Mali Audio \\collection ~\citep{ZANONBOITOETAL_2022NigerMaliAudio} \citep{ZANONBOITOETAL_2022SpeechResourcesTamasheq}\end{tabular}
& \textcolor{rebuttal}{CC BY-NC-ND}
& \begin{tabular}[c]{@{}c@{}}\textbf{38\,332} \\ 18\,546 / 19\,786 / --\end{tabular}  
& \begin{tabular}[c]{@{}c@{}}\textbf{111:01}\\ 52:15 / 58:46 / --\end{tabular}  
& \begin{tabular}[c]{@{}c@{}}\textbf{357} \\ 192 / 165 / --\end{tabular}  
& \begin{tabular}[c]{@{}c@{}}\textbf{10\,s} \\ 10\,s / 10\,s / --\end{tabular}   
& \begin{tabular}[c]{@{}c@{}} Radio broadcasts \end{tabular} \\\midrule
    
\textbf{Extra Large dataset total}  
&
& \begin{tabular}[c]{@{}c@{}}\textbf{2\,669\,869} \\ 1\,125\,901 / 567\,694 / 331\,655 \end{tabular} 
& \begin{tabular}[c]{@{}c@{}}\textbf{14\,548:58} \\ 5\,382:41 / 2\,409:42 / 6\,756:28 \end{tabular} 
& -   
& -  
& -  \\\bottomrule  

\end{tabular}}
 \\\raggedright \scriptsize{*Composed of audio files not included in the CEFC corpus v2.1, 02/2021; **speakers are not uniquely identified.; ***Stats of CFPP2000, MPF and TCOF have changed a bit due to a change in data extraction; License: CC=Creative Commons; NC=non-commercial; BY= Attribution; SA= Share Alike; ND = No Derivative works; CC0 = No Rights Reserved; \textcolor{rebuttal}{User agreement = open for research and NC.}}
    \label{tab:medium_and_small_datasets}
\end{table*}

\noindent\textbf{Small dataset ($\approx$ 1K hours)} is only composed of the MLS corpus for comparison with Wav2Vec2.0~\citep{baevski2020wav2vec} which uses only read English speech. It is also gender-balanced. \\

\noindent\textbf{Medium-clean dataset ($\approx$ 2.7K hours)} contains MLS and EPAC only to enable further investigation on the impact of spontaneous speech on SSL representations. EPAC is a corpus of conversational speech in broadcast news.  \\

\noindent\textbf{Medium dataset ($\approx$ 3K hours)} includes 2,933\,h of speech, from which 1,115\,h is read speech, 1,626\,h broadcast speech, 123\,h spontaneous speech, 38\,h acted telephone dialogues, and 29\,h acted emotional speech. Regarding gender, we collected 1,824\,h  from male speakers, 1,034\,h from female speakers, and 74\,h from unknown gender.  \\

\noindent\textbf{Large dataset ($\approx$ 7.7K hours)}
has 4 additional corpora: MaSS, NCCFr and Voxpopuli (unlabeled + transcribed). It includes 7,739\,h of speech, from which 1,135\,h is read speech, 1,626\,h broadcast speech, 165\,h spontaneous speech, 38\,h acted telephone dialogues, 29\,h acted emotional speech, and 4744\,h professional speech. Except for NCCFr, no information about gender is given in these datasets.  \\

\noindent\textbf{New Extra large dataset ($\approx$ 14K hours)}
has two additional corpora: audiocite.net and Niger-Mali Audio Collection. Audiocite.net includes freely shareable audiobooks of more than 6\,600 hours. 
We created this dataset specifically for the project and section \ref{sec:audiocite} gives details about how it has been acquired. 
The Niger-Mali Audio Collection is data web-crawled from Studio Kalangou and Studio Tamani websites, with the authorization of Fondation Hirondelle. The gender labels were automatically produced by the \texttt{LIUM\_SpkDiarization} tool~\citep{meignier2010lium}.
With these two added datasets, the Extra-large dataset is then composed of read speech (7,834 hours), broadcast speech (1,737\,h), spontaneous speech (165\,h), acted telephone dialogues (38\,h), acted emotional speech (29\,h), and professional speech (4,744\,h).  \\ 

\noindent\textbf{New Gender-specific datasets ($\approx$ 1k hours)} are built using all datasets present in the Large dataset that contain gender information: MLS, Att-Hack, CaFE, CFPP2000, ESLO2, EPAC, GEMEP, PORTMEDIA, TCOF, NCCFr. For EPAC, we keep the totality of female speech~(385\,h), and downsample the male speech to a comparable amount~(413\,h). This results in 1,041\,h of female speech, and 1,006\,h of male speech in the final gender-specific datasets.  \\

\noindent\textbf{Pre-processing for SSL training:} 
Recordings were segmented using time stamps from transcriptions, or cut every 30 seconds when there was no transcription (VoxPopuli \textit{unlabeled}, audiocite.net). When available, we retrieved speaker labels and gender information. Following ~\citep{baevski2020wav2vec}, we removed utterances shorter than 1\,s, and longer than 30\,s. When possible, overlapping speech sentences were also removed. When necessary, audio segments were converted to mono PCM 16\,bits, 16\,kHz. 

\subsection{Audiocite.net Dataset}\label{sec:audiocite}
Audiocite.net is a corpus of read French speech scrapped from the \href{https://www.audiocite.net/}{www.audiocite.net} website in November 2021 thanks to the kind authorization of the website administrator. The website is composed of voluntary work of speakers who explicitly uploaded their content under a \textcolor{rebuttal}{CC licence}
\footnote{\textcolor{rebuttal}{Some of them have supplementary conditions as SA (Share Alike), ND (No Modification) or NC (No commercial Use) while others are in the public domain (CC-0).}} The audiobooks are available online for free and are classified into 15 categories: tales, world news, short stories, poetry, erotic stories, documentaries, science fiction, novels, animals, audiocite-juniors, religions, kitchen, philosophies, history and theatre. All the original texts are either in the public domain or under an open license. 

The Audiocite.net is composed of more than 6600 hours or recordings from 130 speakers and can be found distributed on OpenSLR (\href{https://www.openslr.org/139/}{www.openslr.org/139/}) with the same license as the original work. \textcolor{rebuttal}{The recordings were labeled within 15 categories from junior books to philosophy with the novel category being the majority (about 50\%)}. All the recordings are distributed in their raw format as we downloaded them from audiocite.net (with background music, noise, unexpected speakers, mp3 format, mono or stereo). No pre-processing was applied to the files nor ASR performed on them. We, however, added information of the gender in a `best effort' manner by guessing the gender from the name and checking the voice in case of uncertainty. This information must not be considered as ground truth and is only intended to be used for a rough statistical estimate. \textcolor{rebuttal}{This estimate indicates that female voice represents 34\% of the speech duration.} No attempt to remove speech that could be seen as offensive or sensitive was made. Although the dataset is provided with training, validation and testing partitions, the whole corpus was used for LeBenchmark 14K model training. \textcolor{rebuttal}{Further details about the corpus can be found on the OpenSLR website.}


\section{Building an Open Collection of Pre-trained French SSL Models}
\label{sec:models}

%

\nameprojectnew introduces three novels pre-trained French wav2vec 2.0 to the community based on the Extra Large dataset (i.e. 14,000 hours of speech): 14K-light, 14K-large and 14K-xlarge. More precisely, \nameprojectnew is an open collection of 14 pre-trained SSL models made entirely available on the HuggingFace platform\footnote{\url{https://huggingface.co/LeBenchmark}}. It is worth noticing that the number of released SSL models has doubled from \nameproject to \nameprojectnew as four others have been added for preliminary gender analyses from M. Z. Boito et al. \citep{boito2022study}. The latter four models are not depicted in Table \ref{tab:hyperparamModels}, as they were introduced in \citep{boito2022study}.  In practice, the three new models cover different use cases as the 14K-large and 14K-xlarge are expected to deliver top-notch performance in unconstrained and centralized environments while our 14K-light will bring SSL features to more resource-constrained devices. \textcolor{rebuttal}{We decided to replace the standard base model with an even smaller version (14K-light) as previous works \citep{lebenchmark2} have shown minimal performance improvement for base models when the amount of pre-training data was increasing. The evaluation of small models pre-trained from scratch, however, was a fairly open question. } As of now, these additions represent both the most powerful and parameters-efficient SSL-powered models for the French language. 

\subsection{On the Choice of Wav2vec 2.0}

At the time of \textit{LeBenchmark}, wav2vec 2.0 was the only \textcolor{rebuttal}{state-of-the-art} open-source and available SSL pre-training strategy. \textcolor{rebuttal}{Other alternatives such as MockingJay \citep{liu2020mockingjay} or CPC-based methods \citep{chung19_interspeech} were already invented, but could not guarantee, at least from the original articles, state-of-the-art performance once scaled to hundreds of millions of parameters and many thousands of hours of pre-training data.} Wav2vec 2.0 naturally fitted our requirements as it was also achieving state-of-the-art performance. According to the SUPERB benchmark \citep{tsai2022superb}, the three best-performing pre-training strategies to date are WavLM, HuBERT, and wav2vec 2.0. However, no implementation of WavLM may be found to replicate the pre-training process and the reported results. HuBERT, on the other hand, suffers from a much more complex training process due to the iterative refining of the discrete targets obtained with k-means. \textcolor{rebuttal}{At the time of writing, a new faster implementation of HuBERT is now available \citep{chen2023reducing}, but this was not the case when \textit{LeBenchmark} started}. Furthermore, and as depicted in \citep{tsai2022superb}, the downstream performance of \textit{BASE} and \textit{LARGE} models for HuBERT and wav2vec 2.0 are similar despite a slight advantage for HuBERT, potentially originating from an extensive hyperparameters tuning. Indeed, from our experience, the SSL pre-training behavior varies significantly following hyperparameter changes. In summary, the wav2vec 2.0 architecture enables \nameprojectnew to compare fairly with previously introduced French models while retaining state-of-the-art performance compared to existing alternatives. 

\textcolor{rebuttal}{Finally, and despite a clear interest, it quickly appeared intractable to consider pre-training from scratch multiple SSL methods. Indeed, and as shown in Table \ref{tab:summary}, the total energy cost as well as necessary compute resources would have exceeded by far the available limits. Hence, we decided to focus on a single architecture while giving to the community both the full pre-training dataset as well as downstream tasks to further push the study with better or newly introduced SSL algorithms.}

\begin{table}[!t]
\caption{Summary of the pre-trained wav2vec 2.0 models delivered with LeBenchmark and LeBenchmark 2.0. Newly released models are denoted in bold. ``GPU Hours'' refer to the total training time cumulated over ``GPU Count'' to reach the number of ``Updates''.}
\begin{center}
\resizebox{\textwidth}{!}{
    \begin{tabular}{lccccccc}
    \toprule
        \textbf{Model} & \textbf{Pre-training data} & \textbf{Parameters Count}  & \textbf{Output Dimension} & \textbf{Updates} & \textbf{GPU Count} & \textbf{GPU Hours} \\
    \midrule
           1K-\textit{base}  & 1,096\,h & 90M   & 768  & 200K & 4 & 1,000 \\
           1K-\textit{large}  & 1,096\,h & 330M  & 1024  & 200K & 32 & 3,700 \\
    \midrule
          2.7K-\textit{base}   & 2,773\,h & 90M  & 768  & 500K & 32 & 4,100 \\
    \midrule
          3K-\textit{base}  & 2,933\,h & 90M  & 768  & 500K & 32 & 4,100 \\
          3K-\textit{large}  & 2,933\,h & 330M  & 1024  & 500K & 32 & 10,900 \\
    \midrule
          7K-\textit{base}  & 7,739\,h & 90M  & 768 & 500K & 64 & 7,900 \\
          7K-\textit{large}  & 7,739\,h & 330M  & 1,024 & 500K & 64 & 13,500 \\
    \midrule
          \textbf{\nameprojectnew} &  &   &  &  &  &  \\
          \textbf{14K-\textit{light}} & 14,000\,h & 26M  & 512 & 500K & 32 & 5,000 \\
          \textbf{14K-\textit{large}} & 14,000\,h & 330M  & 1,024 & 1M & 64 & 28,800 \\
          \textbf{14K-\textit{xlarge}} & 14,000\,h & 965M  & 1,280 & 1M & 104 & 54,600  \\
         \bottomrule
    \end{tabular}}
\end{center}
\label{tab:hyperparamModels}
\end{table}

\subsection{Pre-training Hardware and Sofware Environments}

Large-scale SSL pre-training was mostly conducted within the Jean Zay French supercomputer converged platform made available to researchers by GENCI\footnote{GENCI Jean Zay official presentation: \url{http://www.idris.fr/eng/jean-zay/jean-zay-presentation-eng.html}}. As of January 2023, Jean Zay is offering access to 2,696 Nvidia Tesla V100 (16GB and 32GB) split between four and height GPU per node with dual Intel CPU as well as 416 Nvidia Tesla A100 80GB with dual AMD CPU. Jean Zay is mostly powered by nuclear energy, hence facilitating a low carbon emission rate per FLOPS. The reported Power Usage Effectiveness (PUE) ratios are 1.21 and 0.65 depending on the scenario (i.e. considering the heat transfer to nearby infrastructures), putting Jean Zay in the list of the most efficient supercomputers worldwide\footnote{\url{https://systematic-paris-region.org/wp-content/uploads/2022/06/slideshow-Hub-Day-HPC-Hybride.pdf}}. Most models except 
14K-xlarge have been trained on nodes equipped with four 32GB Nivida Tesla V100 hence triggering multi-node training to reach the desired 32 or 64 GPU. 14K-xlarge was trained with 80GB Nvidia Tesla A100 nodes equipped with height GPU each. 
Data read and write operations were made throughout a fast Nested File System (NFS) without any streaming library. A total of 2.9 TB of storage was necessary to hold the entire 14,000 hours dataset. In practice, wav2vec 2.0 models could be trained with much fewer and less powerful GPU, but at the cost of significantly longer training time (i.e. mostly intractable) due to the gradient accumulation needed to reach the large batch size required by the contrastive learning nature of wav2vec 2.0.

The speech and audio processing toolkits landscape has significantly expanded in the last decade, however, only two tools support the full wav2vec 2.0 pre-training: Fairseq \citep{ott2019fairseq} and SpeechBrain \citep{ravanelli2021speechbrain}. In practice, all models trained with the Large dataset (7,000 hours) or smaller sets have been produced with Fairseq, while the others have been trained with SpeechBrain by adapting the wav2vec 2.0 CommonVoice recipe to our data. \textcolor{rebuttal}{The change to SpeechBrain was motivated by a simpler and faster pre-training. In practice, we also re-trained a 7k-large model with SpeechBrain to make sure that the performance was identical to the Fairseq model.} The Python environments are corresponding to those detailed in the toolkit installation scripts attached to each commit.

\subsection{Wav2vec 2.0 Hyperparameters}

The wav2vec 2.0 architecture can be summarized in four distinct blocks: an acoustic feature extractor made of a convolutional neural network, a latent or contextual extractor composed of a Transformer network, a quantization module, and the final contrastive block. The entire detailed list of architectural hyperparameters (i.e. more than 70 parameters) for each model can be found in the corresponding HuggingFace repository\footnote{\url{https://huggingface.co/LeBenchmark}}. In short, all models share the same CNN encoder architecture and mostly differ in the hidden dimension size and depth of the Transformer and quantizer. For instance, the sizes of the intermediate and hidden layers are $[\textcolor{rebuttal}{2048}, 3072, 4096, 5120]$ and $[\textcolor{rebuttal}{512}, 768, 1024, 1280]$ for the \textcolor{rebuttal}{\textit{light}}, \textit{base}, \textit{large}, and \textit{xlarge} models respectively. The number of blocks in the Transformer also increases following the model size with $[6, 12, 24, 48]$. In practice, \nameprojectnew follows the configurations initially reported by A. Baevki et al. \citep{baevski2020wav2vec}, A. Babu et al. \citep{babu2021xls} \textcolor{rebuttal}{and T. Ashihara et al. \citep{ashihara22_interspeech}} as extensive hyperparameter and architecture searches are intractable even with Jean Zay resources.

\subsection{Pre-training Hyperparameters}

The extensive list of pre-training hyperparameters is reported in the \nameproject HuggingFace repository while the most impactful ones are given in the following. The duration of each model pre-training is measured in \textit{``steps''} or \textit{``updates''} referring to an effective update of the neural network weights or a call to the \textit{``.backward()''} function in PyTorch. This quantity varies with the amount of available data as well as the number of neural parameters to optimize. For instance, the 14K-xlarge made one million updates compared to 200,000 for the 1K-large. Increasing the number of updates ultimately leads to better downstream performance. Nevertheless, the latter behavior must be contrasted with the high cost associated with longer training times as many dozens of GPU are being used at once. Again, we fixed the number of steps according to the original wav2vec 2.0 and XLS-R. All models are trained with the Adam optimizer and decoupled weight decay (i.e. AdamW) \citep{loshchilov2018decoupled} following a two-step scheduler made of around $8\%$ of warmup steps and a polynomial decay to zero.

Each training step is then associated with an effective batch size measured, for convenience, in seconds. All models from \nameprojectnew have been trained with an effective batch size of between two and three hours. For instance, the 14K-large model used 40 GPU that could fit 118 seconds of speech each per batch alongside a gradient accumulation factor of two, resulting in a total effective batch size of $(40 \times 118 \times 2)/3600 = 2.63$ hours of speech signal per step. 

Ensuring a constant effective batch size necessitates a constant amount of signal per GPU. To this extent, both Fairseq and SpeechBrain toolkits implement a dynamic batching mechanism that automatically bundles samples together depending on their size to match the desired maximum boundary. The latter boundary depends on the VRAM capacity of the GPU and varies with the size of the model. For instance, the 14K-large model was trained with a boundary of 118 seconds on 32GB GPU while the 14K-xlarge model stayed at 60 seconds with 80GB GPU.

To limit the VRAM consumption due to the quadratic increase in complexity of the Transformer self-attention following the sequence length, all samples are cropped at 20 seconds. The remaining signal is simply used as a different example. Similarly, and to avoid masking the entirety of the sample for contrastive loss, segments shorter than 2 seconds are removed from Fairseq models while they are concatenated with SpeechBrain. For the largest models, i.e. 14K-xlarge, audio samples are cropped at 15 seconds as no real degradation from slightly shorter sequences is to be expected, as demonstrated by Y. Gao et al \citep{gao2022match}.

Finally, masks applied to the output of the feature extractor (i.e. CNN) are of 10 consecutive frames for all models. Masking probabilities, however, change with the model size. Indeed, $50\%$ of the frames are masked for \textit{base} models compared to $75\%$ for the \textit{large} and \textit{xlarge} models \textcolor{rebuttal}{ as reported by A. Baevki et al. \citep{baevski2020wav2vec}, A. Babu et al. \citep{babu2021xls}}.

\subsection{Wav2vec 2.0 Pre-training: tips and tricks}

Due to the very high entry ticket to pre-training large SSL models, almost no resources exist to help the community with this process. In particular, we found both Fairseq and SpeechBrain wav2vec 2.0 pre-training recipes do not simply transfer seamlessly to the \nameprojectnew datasets. Such difficulties in training certainly originate from the high complexity of the pipeline: hundreds of millions of neural parameters on thousands of hours of speech with dozens of distributed GPU. In the following, we propose a few tips and tricks to avoid the two most common issues encountered while pre-training \nameprojectnew models: exploding losses and collapsing representations. 

Exploding losses (i.e. NaN values) were the most common issue faced when training for longer periods of time. Indeed, all models trained for more than 500M steps experienced infinite losses at some point whether it was with Fairseq or SpeechBrain. As expected, simply changing the learning rate did not help as both toolkits already carefully designed the scheduler as well as gradient clipping strategies to avoid any perturbation in the gradient flow. In fact, the mixed precision training was most of the time responsible for this issue. Hence, upon explosion, a simple resuming of the training with full precision (i.e. fp32) was sufficient to finish the training. \textcolor{rebuttal}{The resulting VRAM increase was low enough to enable training with the same batch size.} This may be explained by extremely small weights or values appearing after extended training with AdamW due to the weight decay, hence reaching the rounding-to-zero floor of 16-bit floats. It is worth noticing that switching to bfloat16 (i.e. fp32 values range) instead of float16 solved entirely the issue with SpeechBrain while preserving the training speed. 

Collapsing representations may happen when the contrastive task is too hard or too simple. It may easily be spotted at training time as the accuracy, defined as the cosine similarity between projected and quantized latent representations over the initially masked frames, quickly jumps to values close to $100\%$. In practice, this can easily be avoided by increasing the diversity loss and the mask length or probability. Indeed, a very high accuracy may simply mean that only very few entries of the quantization codebook are used, hence easy to predict. The latter phenomenon may especially arise with a dataset composed of short audio segments.

\section{Standardized and Replicable Benchmark of French Tasks for SSL Models}
\label{sec:benchmark}

%
%


\nameprojectnew introduces two new candidates to \nameproject for a total of six tasks: automatic speech recognition (ASR), spoken language understanding (SLU), automatic speech translation (AST), automatic emotion recognition (AER), syntactic analysis (SA) and automatic speaker verification (ASV). We designed our benchmark to best cover the different criteria that the community may expect from pre-trained extractors. In particular, SSL models must integrate information related to transcription (ASR), semantics (SLU, SA), translation (AST), and paralinguistics (AER, ASV). To validate the robustness of SSL to varying data volumes, we also selected corpora matching all potential use cases: high (ASR\textcolor{rebuttal}{, AST}), medium (SLU, AST, ASV), and low (\textcolor{rebuttal}{ASR, }AER, SA) resource. \\

\noindent\textbf{SSL Models configurations.} In all the following evaluations, SSL models may be described as \textit{``fine-tuned''} or \textit{``frozen''} and \textit{``task-agnostic''} or \textit{``task-specific''}. Indeed, and contrary to the SUPERB benchmark, we are investigating different scenarios corresponding to various real-life applications. In SUPERB, all models are frozen, meaning that the neural parameters of the SSL models are not updated with the downstream task of interest. Having a heterogeneous set of downstream decoders is of crucial importance to avoid variations in the final ranking as demonstrated by S. Zaiem et al. \citep{zaiem2023benchmark}. In \nameprojectnew, we also investigate the results obtained with fine-tuned models, where both the SSL model and the downstream decoder are trained to solve a given task, hence reaching much higher levels of performance. The latter is done at the cost of a more compute resource-intensive fine-tuning stage. Task-agnosticism or specificity simply defines the standard pre-trained SSL models or their already finetuned equivalent. For instance, one may wish to first fine-tune a wav2vec 2.0 architecture on speech recognition before evaluating it on SLU. The latter ASR-specific model is referred to as being ``task-specific''. \\

\noindent\textbf{Considered SSL baselines.} Among the different evaluations reported below, \nameprojectnew aims at providing two different levels of study: (a) evaluate the relative performance improvement between the different provided French SSL models, and (b) evaluate the relevance of language-specific SSL models against multilingual or out-of-domain large-scale models. First, and as \nameprojectnew is the only resource available for French SSL, the former comparison will be only conducted with our own models. Second, \nameprojectnew wav2vec will be compared to \textcolor{rebuttal}{XLSR-53~\citep{conneau21_interspeech}} 
and XLS-R-1B~\citep{babu2021xls}\textcolor{rebuttal}{, which for consistency we refer as \textit{XLS-R-xlarge},} for the multilingual aspect. 
\textcolor{rebuttal}{Whisper~\citep{radford2022robust} will not be considered as two major drawbacks prevent its use in a fair comparison: (a) the training data is virtually unknown and may already contain the train or test sets of our downstream tasks, and (b) it is based on weakly supervised training, not SSL. }
\subsection{Automatic Speech Recognition}
\label{subsec:asr}

%
%
%

Automatic speech recognition is a common downstream task to evaluate SSL models. We investigated the behavior of the different \nameprojectnew models with two scenarios: a challenging low-resource scenario with only 22~hours of training data on TV and radio shows, and a high-resource scenario with 428 hours of read speech. \\

%
%
%
%

\noindent\textbf{Downstream datasets.} Two different French datasets were used.
The first one is the ETAPE corpus. This is the official dataset released during the French ETAPE evaluation campaign in 2011~\citep{gravier2012etape}. It is composed of diverse French shows: TV news, TV debates, TV amusement, and radio shows. These shows are split into three subcorpora -- training: 22\,h, validation: 7\,h, and testing: 7\,h. ETAPE is distributed in the ELRA catalogue.\footnote{\url{https://catalogue.elra.info/en-us/repository/browse/ELRA-E0046/}} It is free for academic research. The second dataset is the French part, version 6.1, of the well-known CommonVoice project~\citep{ardila2020common}. This project started in July 2017 and employs crowdsourcing for both speech data collection and human validation for a large number of languages. 
The speech data is made of read sentences extracted from Wikipedia. It contains 428\,h, 24\,h, and 25\,h of speech data for the training, validation, and testing sets respectively. \\

%
%
%
%

\noindent\textbf{Downstream models and hyperparameters.} To conduct our experiments, we employed the SpeechBrain toolkit~\citep{ravanelli2021speechbrain}, which is built on PyTorch and designed specifically for speech-processing tasks. 
Additionally, we utilized the Hugging Face version of the \nameproject models even though fairseq checkpoints are also available. The SpeechBrain toolkit offers a diverse array of recipes, and to ensure the reproducibility of our experiments, we followed the SpeechBrain recipe specific to the CommonVoice ASR task. In both of the aforementioned scenarios, we initiated with a pre-trained \nameproject model and added three additional dense hidden layers of size 1,024 on top with random initialization. Each of these layers was associated with the LeakyReLU activation function. Subsequently, we performed fine-tuning for speech recognition on the training data by applying the SpecAugment data augmentation technique. For optimization, we employed the Connection Temporal Classification (CTC) loss function~\citep{Graves:2006:CTC:1143844.1143891}. The overall model's output consists of 78 tokens, encompassing both characters, sub-word units, and the CTC blank character. This number is higher than in English due to the presence of numerous accented letters in French. For example, variations derived from the letter \textsl{e} include \textsl{é}, \textsl{è}, \textsl{ê}, and \textsl{ë}. Other letters like \textsl{ç} or \textsl{œ} have also to be taken into account. Following the SpeechBrain recipe for CommonVoice, we optimized the model using two optimizers. The Adam optimizer was dedicated to the parameters derived from the \nameproject wav2vec2.0 model, while the AdaDelta optimizer was used for all the parameters on top of it. We applied a dropout technique to train the three top dense hidden layers, with a probability of 0.15. For each experiment, the training was made on 80 epochs, keeping the model that reached the best results on the development data.\\

\noindent\textbf{LeBenchmark 1.0. }\textcolor{rebuttal}{In a prior study \citep{lebenchmark2}, we demonstrated that \nameproject SSL models pre-trained solely on French data outperformed comparable models pre-trained on English~(e.g. wav2vec~2.0 on Librispeech) or multilingual data~(e.g. XLSR-53), which also included French. Our new findings focus on assessing the impact of additional data on the pretraining of the French dataset used for SSL. Moreover, we exclude our previous frozen SSL setting as it is consistently inferior to end-to-end fine-tuning, and we find that it is also a setup that is becoming less and less relevant to the ASR community.
Our previous investigations also showed that end-to-end ASR fine-tuning is consistently more effective using large models as initialization, and thus our \nameproject benchmark for ASR present results only for large versions of already evaluated models~(1K, 3K, 7K), and for all new models~(14K) from Table~\ref{tab:hyperparamModels}. Finally, for baselines, we consider XLS-R-xlarge only, as XLSR-53 was previously evaluated, and it presented inferior results to all \nameproject models (base or large).} 

%
%
%
%

\noindent\textbf{Results analysis and discussions.} 
As depicted in Table~\ref{tab:res_asr_sb_models}, and consistent with the results described in~\citep{lebenchmark2}, the \oL model exhibits a higher word error rate compared to the \tL model.
Section~\ref{subsec:overviewpretrainingdatasets} provides detailed information about the content of the pretraining datasets used for the different models.
Incorporating 2,000 hours of primarily broadcast news speech with the initial 1,000 hours of read speech used to pretrain the 1K-large model significantly improves the performance of the 3K-large model for speech recognition.
However, further augmentation with 7,000 hours of formal read or prepared speech (parliamentary events) does not yield substantial improvement for the 7K-large model.
Nevertheless, the performance of the 7K-large model is still significantly better than the 3K-large on broadcast news (ETAPE) and comparable to read speech (CommonVoice). This phenomenon is worsened by the introduction of the 14K hours dataset, as 14K-large and 14K-light are not able to outperform even the 3K-large. This can be explained by the nature of the added data, i.e. read speech, that may simply not help at reaching better performance above a certain threshold. In most cases, the 14K-light model exhibits degraded performance for automatic speech recognition even though speech recognition rates remain better than those of XLSR-53 reported in \citep{lebenchmark2}. 

\begin{table}
  \caption{ASR results in terms of word error rate (WER\%, lower is better) on Common Voice and ETAPE corpora, with pre-trained wav2vec~2.0 models fine-tuned on labeled ASR data. {\color{gray}Gray numbers} denote the standard deviation.} 
  \footnotesize
  \label{tab:res_asr_sb_models}
  \centering
  \scalebox{1}{
  \begin{tabular}{l|cc|cc}
    \toprule
        {\textbf{Corpus}} & \multicolumn{2}{c|}{\textbf{CommonVoice}} & \multicolumn{2}{c}{\textbf{ETAPE}} 
        \\ \midrule
        \textbf{Representation}	 &	\textbf{Dev} &	\textbf{Test} & \textbf{Dev} & \textbf{Test}\\  \midrule
1K-\textit{large} &
9.49{\footnotesize\color{gray}$\pm0.20$} & 
11.21{\footnotesize\color{gray}$\pm0.23$} &
28.57{\footnotesize\color{gray}$\pm0.79$} &
30.58{\footnotesize\color{gray}$\pm0.88$}\\  
3K-\textit{large} &
\textbf{8.00}{\footnotesize\color{gray}$\pm0.19$}& \textbf{9.27}{\footnotesize\color{gray}$\pm0.20$} &
22.26{\footnotesize\color{gray}$\pm0.76$}& 24.21{\footnotesize\color{gray}$\pm0.85$}\\ 
7K-\textit{large}  &
8.02{\footnotesize\color{gray}$\pm0.18$} & 9.39{\footnotesize\color{gray}$\pm0.21$} &
\textbf{21.34}{\footnotesize\color{gray}$\pm0.74$}& \textbf{23.46}{\footnotesize\color{gray}$\pm0.83$}\\ 
\midrule
14K-\textit{light}  &
19.86{\footnotesize\color{gray}$\pm0.28$} & 22.81{\footnotesize\color{gray}$\pm0.34$} &
58.30{\footnotesize\color{gray}$\pm0.66$}& 59.82{\footnotesize\color{gray}$\pm0.7$}\\ 
14K-\textit{large}  &
8.39{\footnotesize\color{gray}$\pm0.19$} & 9.83{\footnotesize\color{gray}$\pm0.21$} &
23.67{\footnotesize\color{gray}$\pm0.81$}& 26.03{\footnotesize\color{gray}$\pm0.89$}\\ 
14K-\textit{xlarge} & 
8.26{\footnotesize\color{gray}$\pm0.19$} & 9.83{\footnotesize\color{gray}$\pm0.21$} &
22.38{\footnotesize\color{gray}$\pm0.95$}& 24.67{\footnotesize\color{gray}$\pm0.83$}\\
    \bottomrule
  \end{tabular} }
\end{table}
\subsection{Spoken Language Understanding}
\label{subsec:slu}

%
%
%

Spoken Language Understanding (SLU) aims at extracting semantic representations from speech signals containing sentences in natural language \citep{DeMori1997:SDBook}.
Classical approaches to SLU used a cascade model made of an 
\textcolor{rebuttal}{ASR} system feeding a Natural Language Understanding (NLU) module \citep{Raymond2006:FST-SEM,dinarelli-etal-2009-ranking,dinarelli09:Interspeech,Quarteroni.etAl:Interspeech09,Hahn.etAL-SLUJournal-2010,caubriere:hal-02465899,ghannay:hal-03372494}.
Neural networks led to large advances for \emph{end-to-end} SLU systems \citep{10.1007/978-3-319-77113-7_4,DBLP:journals/corr/abs-1802-08395,dinarelli:hal-01553830,desot:hal-02464393,lugosch2019speech,DBLP:journals/corr/abs-1906-07601,dinarelli2020data,pelloin:hal-03128163}, 
which are preferred to cascade systems, in particular for their ability to reduce error propagation effects and to exploit acoustic components to deduct semantic information \citep{DESOT2022101369}. \\

%
%
%
%

\noindent\textbf{Downstream datasets.} For French SLU benchmarking we used the well-known MEDIA corpus \citep{bonneau-maynard-etal-2006-results}, used also in \citep{lebenchmark2} and allowing thus for direct comparison. The MEDIA corpus focuses on the domain of hotel information and reservations in France.
It is made of 1,250 human-machine dialogues transcribed and annotated with 76 semantic concepts.
The complete MEDIA task requires models to extract both concepts and their normalized values. For instance from the (chunk of speech whose transcription is the) phrase \emph{``two double rooms''} the model must predict the following semantic annotation: \texttt{room-number}[2] \texttt{room-type}[double]. All models applied to this task in the past were trained to extract concepts only (\texttt{room-number} and \texttt{room-type}), while normalized values are extracted in a post-processing step using rule-based modules. The latter is necessary because of the presence of open-domain values, like dates or amounts (e.g. \emph{``from October 26th ...''} must be annotated as \texttt{start-date[26/10/2024]}), which are difficult to normalize for models trained from scratch, while that is an easy task for rule systems.
Like in previous work thus \citep{lebenchmark1,lebenchmark2}, we train models to predict tokens, concept boundaries and concepts directly from speech. This means models must learn to transcribe, segment and annotate speech signals, which is a much more difficult task than predicting domain, intent and slot-value pairs like in more traditional SLU tasks such as FSC. A comparison of performances on MEDIA and FSC tasks can be appreciated in our previous work~\citep{dinarelli:hal-03872546}, although results on FSC are always given in terms of accuracy, even an end-to-end model with basic features reaches an accuracy over 90\%.

The MEDIA corpus is split into 12,908 utterances (41.5 hours of speech) for training, 1,259 for development (3.5 hours), and 3,005 for test (11.3 hours). \\

%
%
%
%

\noindent\textbf{Downstream models and hyperparameters.} SLU models used in this paper are the same as in \citep{lebenchmark2}, few modifications have been introduced which will be described along this section.
Such models have a \emph{sequence-to-sequence} architecture based on LSTMs and attention mechanisms \citep{Hochreiter-1997-LSTM,DBLP:journals/corr/BahdanauCB14}.
The encoder has a similar pyramidal structure as the one proposed in \citep{DBLP:journals/corr/ChanJLV15}, the decoder uses two attention mechanisms, one for attending the encoder's hidden states, and one for attending the decoder's previous predictions, like the self-attention module of Transformers \citep{NIPS2017_3f5ee243}.
One difference with respect to LeBenchmark models proposed in \citep{lebenchmark2} is that we added a layer normalization after each decoder layer, which made learning more stable when using features extracted with SSL models as input, and improved the model's generalization.
All models were trained to minimize the CTC loss \citep{Graves:2006:CTC:1143844.1143891}. We note that our models can be trained also with a \emph{cross-entropy} loss, which is often used with sequence-to-sequence architectures predicting discrete symbols. However in preliminary experiments we found that CTC loss leads always to significantly better performances.
In all our experiments we use SSL models as feature extractors. Features were given as input to SLU models as an alternative to traditional features (e.g. MFCC). Following \citep{lebenchmark2}, we used both task-agnostic and task-specific SSL models. In the task-specific case, SSL models were fine-tuned for the ASR output like in \citep{lebenchmark2}, and we use XLSR-53-large and XLS-R-xlarge as baselines.

Models described in \citep{lebenchmark2} were learned with three training steps. Each training step uses the model learned in the previous step for initializing the current model's parameters.
While this strategy is the most effective, it implies a relatively high training cost.
In this work we instead use full end-to-end training such as in \citep{dinarelli:hal-03872546}. Hence, models are learned from scratch in a single training procedure.
In addition, in this work, we tested a multi-task learning setting where the encoder and the decoder are learned with two different CTC losses: with respect to the ASR output for the encoder; with respect to the SLU output for the decoder following a standardized output format \citep{lebenchmark2}. This multi-task learning setting will be indicated with \textit{mt} in Table~\ref{tab:SLU-MEDIA-results}.
In order to study the impact of the SLU model size on results, especially when using features from SSL models, we tested hidden layers of size 256 and 300. Since these gave similar results in most cases, we did not further optimize the model size.
We also found it benef7icial, when using fine-tuned SSL model's features, to increase the temperature at the output softmax layer to 2 (indicated as \emph{t2} in the table). This strategy has been used successfully for model distillation \citep{DBLP:journals/corr/abs-1910-01108}, and intuitively has a similar effect as using smoothed label representations as targets \citep{10.5555/3454287.3454709}. Beyond these differences, all models use the same hyperparameters as those used in \citep{lebenchmark2}. \\

\begin{table}[!t]
\caption{End2End SLU results in concept error rate (CER \%, lower is better $\shpos$) on the MEDIA corpus. ``h256'' and ``h300'' refer to a hidden size of 256 and 300 and neurons respectively, while ``mt'' is a multi-task ASR-SLU training. ``t2'' means that a Softmax temperature of value 2 was applied.}
\scriptsize
\begin{center}
\scalebox{1}{
    \begin{tabular}{l|c|c|c|c|c}

        \toprule
        \multicolumn{6}{c}{Corpus: MEDIA, Metric: Concept Error Rate (CER \%) $\shpos$} \\
        \midrule
        \textbf{Representation} & \textbf{Model} & \textbf{Params (SSL/SLU)} &  \textbf{RawER} & \textbf{Dev} & \textbf{Test} \\
        \midrule
        \multicolumn{6}{c}{\textbf{Task-agnostic models}} \\
        \hline
         spectrogram & h300 &   -/13.18 &  59.36 &    64.96  &   58.12 \\
         spectrogram & h300 mt &   -/13.18 &  \textbf{30.44} &    30.24 &   \textbf{30.39} \\
        \midrule
         7K-\textit{large} & h256 \citep{lebenchmark2} &   330/12.18 &  - &    19.68  &   18.77 \\
         7K-\textit{large} & h300 &   330/15.45 &  17.26 &    18.57  &   16.99 \\
         7K-\textit{large} &h300 mt &   330/15.45 &  \textbf{15.36} &    \textbf{16.62}  &   \textbf{15.47} \\
        \hline
         14K-\textit{light} & h256 &   26/12.18 &  22.71 & 22.62 & 20.92 \\
         14K-\textit{light} & h256 mt &   26/12.18 &  \textbf{19.29} & \textbf{19.41} & \textbf{18.67} \\
        \hline
         14K-\textit{large} & h300 &   330/15.45 &  18.26 & 18.63 & 17.35 \\
         14K-\textit{large} & h300 mt &   330/15.45 &  \textbf{15.91} & \textbf{16.62} & \textbf{14.43} \\
        \hline
         14K-\textit{xlarge} & h256 &    965/12.70 &  21.44 & 17.52 & 16.24 \\
         14K-\textit{xlarge} & h256 mt &    965/12.70 &  \textbf{14.73} & \textbf{15.66} & \textbf{14.43} \\
        \midrule
         XLSR-53-\textit{large} & h256 \citep{lebenchmark2} &    330/12.18 &  - & \textbf{18.45} & 18.78 \\
         XLSR-53-\textit{large} & h256 &    330/12.18 &  18.38 & 18.99 & 18.68 \\
         XLSR-53-\textit{large} & h256 mt &    330/12.18 &  \textbf{17.74} & 18.84 & \textbf{17.16} \\
        \hline
         XLS-R-\textit{xlarge} & h300 &    965/16.07 &  18.02 & 20.75 & 30.08 \\
         XLS-R-\textit{xlarge} & h300 mt &    965/16.07 &  18.53 & \textbf{18.57} & \textbf{29.07} \\
        \midrule
        \multicolumn{6}{c}{\textbf{Task-specific models (ASR fine-tuning)}} \\
        \hline
         7K-\textit{large} & h256 \citep{lebenchmark2}           &   330/12.18 &  - & 14.58 & 13.78 \\
         7K-\textit{large} & h300 t2          &   330/15.45 &  10.82 & \textbf{13.53} & 12.80 \\ 
         7K-\textit{large} & h300 mt t2           &   330/15.45 &  10.83 & 13.95 & \textbf{12.71} \\
        \hline
         14K-\textit{light} & h300           &   26/15.45 &  15.07 & 17.03 & 20.02 \\
         14K-\textit{light} & h300 mt          &   26/15.45 &  \textbf{13.66} & \textbf{15.48} & \textbf{18.22} \\
        \hline
          14K-\textit{large} &  h256 t2           &   330/12.18 &\textbf{10.43} & \textbf{12.96} & 13.78 \\
         14K-\textit{large} & h256 mt t2          &   330/12.18 &  13.76 & 14.43 & \textbf{12.85} \\
        \hline
         14K-\textit{xlarge} & h256 t2           &   965/12.70 &  \textbf{11.19} & \textbf{13.74} & \textbf{13.88} \\
          14K-\textit{xlarge} & h256 mt t2          &   965/12.70 & 11.35 & 14.58 & 14.53 \\
        \midrule
         XLSR-53-\textit{large} & h300          &   330/15.45 &  18.45 & 17.42 & \textbf{15.01} \\
         XLSR-53-\textit{large} & h300 mt         &   330/15.45 &  \textbf{13.09} & \textbf{15.22} & 16.60 \\
        \hline
         XLS-R-\textit{xlarge} & h300 t2          &   965/16.07 &  14.87 & 17.22 & 24.15 \\
         XLS-R-\textit{xlarge} & h300 mt t2          &   965/16.07 &  \textbf{13.39} & \textbf{15.42} & \textbf{23.30} \\
        \hline
        
    \end{tabular}
}
\end{center}
\label{tab:SLU-MEDIA-results}
\end{table}

%
%
%
%
\noindent\textbf{LeBenchmark 1.0.} \textcolor{rebuttal}{Similarly to ASR, in the previous version of \nameproject we also observed consistent inferior performance of features extracted from base models, compared to their large counterparts. As we do not expect our differences in training regime and model architecture to impact this conclusion, in this work we present complementary results only for the 7K-large, the best of the \emph{LeBenchmark-1.0} models on SLU, and for all new models~(14K) from Table~\ref{tab:hyperparamModels}.  Results with the spectrograms~(our basic features), 7K-large, XLSR-53-large features are comparable to \citep{lebenchmark2}. The other results are contributions of this work. Results with 7K-large allow thus to distinguish improvements due to architecture modifications (see paragraph \emph{Downstream models and hyperparameters}) from improvements coming from new SSL models.}
\\

\noindent\textbf{Results analysis and discussion.} All results on the SLU task are depicted in table~\ref{tab:SLU-MEDIA-results}. We report Concept Error Rate (CER, the lower the better) on development and test data (respectively columns \textbf{Dev} and \textbf{Test} in the table), as well as the raw error rate (column \textbf{RawER}) on the development data, which is the error rate computed on the raw output of the system. The raw output of the system includes words, concept boundaries and concepts, please refer to the appendix of https://aclanthology.org/2020.wmt-1.71/\citep{lebenchmark2} for details. CER is computed on concepts only.
For each experiment, we report the best results with respect to the hidden layer size on the development data, and its corresponding multi-task learning setting (\emph{mt}). We also specify \emph{t2} in the table if the best results were obtained with a temperature of 2 at the output softmax.

Our best result on validation data with spectrogram features is 30.44, which is only slightly worse than the 29.07 obtained in \citep{lebenchmark2}, with the advantage that in this work, the model is trained end-to-end in one step with the multi-task setting. Additionally, the increased generalization power of the model allows us to reach an error rate of 30.39 on test data, which is slightly better than the 31.10 reported in \citep{lebenchmark2}.

Using input features from \nameproject models (7K-large; 14K-light, large, and xlarge) improvements on SLU results are impressive, with the best CER respectively on validation and test data of 15.66 and 14.43 obtained with the French wav2vec2 14K-xlarge model's features.
It is interesting to see, yet expected, that the more data is used to train the SSL model, and the bigger the SSL model in terms of parameters, the better the SLU results are. This trend is not completely respected with task-specific fine-tuned models, in particular with the 14K-large model. Most intuitively, this is because of the small amount of data available for the downstream task, which does not allow for an optimal fine-tuning of so many parameters. The fact that SSL models are tuned for ASR output may also play a role since SLU output already contains tokens instantiating concepts and this may lead the model toward more accurate token predictions which do not always correspond to tokens triggering concepts. This last fact is supported by raw error rates (\textbf{RawER} column), accounting for tokens, concept boundaries, and concepts, where not always the best result corresponds to the best CER on validation or test data.
On an absolute scale nevertheless, results obtained with fine-tuned French wav2vec2 models are the best. The concept error rate of 12.71 on test data, obtained with the \emph{7K-large} model, is the new state-of-the-art on this task with an end-to-end model. When using fine-tuned model features, the best results on the test data do not always correspond to the best results on validation data, underlying that probably a more accurate hyper-parameter tuning is needed.
An interesting outcome of SSL fine-tuned models is that the best results are almost always obtained with an increased softmax temperature (\emph{t2}). In particular, we observed erratic training behaviors with the default temperature, leading often to gradient explosions. Using an increased temperature not only allows us to obtain good results but also stabilizes model training.

For comparison, we also experimented with some multi-lingual models from the literature, namely XLSR-53-large and XLS-R-xlarge \citep{babu2021xls}.
While the XLSR-53 large model, both without and with fine-tuning, provides interesting results considering that it is not a model specialized for French, the extra-large model provides clearly inferior performances on the test data compared to French models.
For the XLS-R extra-large model we hypothesize that the larger size of the model together with its multilingualism does not allow a good generalization on a specific French task like SLU.

\subsection{Automatic Speech Translation}
\label{subsec:st}

%
%
%
Automatic speech-to-text translation (AST) consists of translating a speech utterance in a source language to a text in a target language. In this work, we are interested in translating directly from French speech into text in another language, without the use of transcriptions. We investigate two downstream applications for \nameproject models: \textit{hybrid} models and \textit{end-to-end} fine-tuning. For the former, the pre-trained model is leveraged as a feature extractor i.e. frozen. For the latter, \textcolor{rebuttal}{a decoder is} 
appended to the pre-trained model, and the whole architecture is fine-tuned on the target dataset. Training an end-to-end AST model from a pre-trained speech encoder was first proposed in~\citep{li-etal-2021-multilingual}. \\

%
%
%
%

\noindent\textbf{Downstream datasets.} For both AST downstream strategies, we use the multilingual TEDx dataset~\citep{elizabeth2021multilingual}. It covers translation directions from French to three target languages: English~({en}), Spanish~({es}), and Portuguese~({pt}), with following  training sizes: 50\,h~({en}), 38\,h~({es}), and 25\,h~({pt}). For end-to-end fine-tuning, we also present results for the CoVoST dataset~V2~\citep{wang2020covost} containing 302 hours of French speech from CommonVoice version 4.0 translated to English. \\

%
%
%
%

\noindent\textbf{Hybrid downstream models and hyperparameters.} In this set of experiments, we focus on leveraging the pre-trained models as feature extractors, using their output speech representation as input for an end-to-end AST model which is trained from randomly initialized parameters. Inspired by~\citep{nguyen2020investigating,lebenchmark1}, this AST model is an encoder-decoder architecture which takes SSL features as input, passing them through a block of Linear-ReLU followed by 2 convolutional layers with strides of $[2,2]$ and kernel sizes of $[5,5]$. These 1D-convolutional layers reduce the sequence length by $4$ which is then sent to a Transformer~\citep{NIPS2017_3f5ee243} model having 6 layers of encoder, 3 layers of decoder, and hidden dimension $D=256$. This is inspired by the \texttt{s2t\_transformer\_xs} recipe from the fairseq s2t toolkit~\citep{wang2020fairseq}.
For each language pair, we train in total $13$ end-to-end models which take as input features extracted from different SSL pre-trained models shown in Table~\ref{ast:tab:extractionmtedx}. 
We normalize the punctuation of the text before building a $1K$ unigram vocabulary using \texttt{Sentencepiece}~\citep{kudo-richardson-2018-sentencepiece} without pre-tokenization. For GPU efficiency, utterances having more than $3,000$ frames are filtered out. Each of these AST models is trained for $500$ epochs. For all our experiments, we exploit the Adam optimizer~\citep{DBLP:journals/corr/KingmaB14} whose initial learning rate is set to $2e-3$. This learning rate is linearly increased for the first $10K$ warm-up steps then decreased proportionally to the inverse square root of the step counter. The last $10$ checkpoints are averaged and used for decoding with a beam size of $5$. Table~\ref{ast:tab:extractionmtedx} reports the detokenized case-sensitive BLEU computed using \texttt{sacreBLEU}~\citep{post-2018-call}.  \\

\noindent\textbf{End-to-end downstream models and hyperparameters.} End-to-end AST models are trained on SpeechBrain~\citep{ravanelli2021speechbrain} using the HuggingFace Transformers~\citep{wolf-etal-2020-transformers} wav2vec~2.0 interface with spectogram augmentation enabled~\textcolor{rebuttal}{(same setting than Section~\ref{subsec:asr})}. The encoder stack is made of a wav2vec~2.0 model, followed by a linear projection of output dimension 512. The decoder stack is an 8-heads, 6-layers Transformer with feed forward projections of 2,048 neurons and an embedding size of 512. The weights for the wav2vec~2.0 model are initialized from one of the models in Table~\ref{tab:hyperparamModels}, and the model is trained with NLL loss. As for end-to-end ASR models~(Section~\ref{subsec:asr}), two different instances of the Adam optimizer manage the weight updates: one dedicated to the wav2vec~2.0 module, the other one to the following layers. The learning rates are respectively $1e-5$ and $1e-4$. 
The models are trained on a single A100 80GB Nvidia GPU, for 50~(CoVoST), or 
100 epochs~(mTEDx). In all cases, sentences longer than 35\,s are removed for GPU efficiency. For models trained on the mTEDx dataset, 
we found that it was beneficial for performance to remove layer-dropout and dropout within the wav2vec~2.0 stack during training. We hypothesize that this is due to the limited amount of data available for fine-tuning, as large architectures seemed to benefit the most from this modification. The total number of trainable parameters depends on the wav2vec~2.0 model used: it varies between 121.3M~(base), 342.5M~(large), and 989.7M~(xlarge). 
Pre-tokenization strategy is the same as the Hybrid AST setup. 
Lastly, we do not use pre-trained weights to initialize the AST decoder,  \textcolor{rebuttal}{and we do not explore partially freezing the wav2vec~2.0 Transformer encoder layers~(i.e. training only a subset of the layers) as in \citet{boito-etal-2022-trac}.}\\

\noindent\textbf{LeBenchmark 1.0.} \textcolor{rebuttal}{During the first version of this benchmark, only the hybrid setup~(SSL as feature extractor) was explored. In contrast to that, we now also include end-to-end AST fine-tuning, which has since become popular in the speech community. In order to provide the reader with the full context, and allow easy comparison between hybrid and end-to-end approaches, we present results for all models, previously evaluated and new.}\\

\begin{table*}[]
\caption{AST BLEU results (higher is better) of the feature extraction experiments (Hybrid with frozen SSL encoders) on the mTEDx dataset. The best results are in \textbf{bold}. {\color{gray}Gray numbers} denote the standard deviation computed using bootstrap re-sampling~\citep{koehn2004statistical}.}
\centering
\begin{tabular}{lcc|cc|cc}\toprule
\multicolumn{1}{l}{}                                       & \multicolumn{2}{c}{\textbf{fr-en}} & \multicolumn{2}{c}{\textbf{fr-es}} & \multicolumn{2}{c}{\textbf{fr-pt}} \\
\multicolumn{1}{c}{\textbf{Representation}} & \textbf{Dev}   & \textbf{Test}   & \textbf{Dev}   & \textbf{Test}   & \textbf{Dev}   & \textbf{Test}   \\\midrule
1K-\textit{base}~\citep{lebenchmark2}       & 9.18\stdev{0.36}             & 8.98\stdev{0.36}            & 5.09\stdev{0.27}             & 5.64\stdev{0.30}            & 0.39\stdev{0.05}              & 0.49\stdev{0.08}             \\
1K-\textit{large}~\citep{lebenchmark2}       & 15.31\stdev{0.46}             & 14.46\stdev{0.46}            & 13.74\stdev{0.43}           & 14.77\stdev{0.46}            & 8.29\stdev{0.34}              &  9.37\stdev{0.38}          \\\hline
2.7K-\textit{base}~\citep{lebenchmark2}              & 15.09\stdev{0.49}             & 14.69\stdev{0.48}             & 13.27\stdev{0.43}              & 14.04\stdev{0.43}            & 4.72\stdev{0.27}             & 5.51\stdev{0.28}            \\\hline
3K-\textit{base}~\citep{lebenchmark2}                  & 15.05\stdev{0.49}              & 14.80\stdev{0.47}            & 13.19\stdev{0.44}              & 14.27\stdev{0.44}            & 4.44\stdev{0.29}              & 4.72\stdev{0.25}           \\
3K-\textit{large}~\citep{lebenchmark2}                & 17.94\stdev{0.51}             & 18.00\stdev{0.51}            & 16.40\stdev{0.49}             & 18.12\stdev{0.48}            & 8.64\stdev{0.34}              & 9.55\stdev{0.36}             \\\hline
7K-\textit{base}~\citep{lebenchmark2}        & 15.13\stdev{0.45}              & 14.50\stdev{0.45}             & 12.78\stdev{0.40}             & 13.61\stdev{0.44}            & 2.65\stdev{0.20}             & 2.66\stdev{0.23}            \\
7K-\textit{large}~\citep{lebenchmark2}                 & \textbf{19.23}\stdev{0.54}             & \textbf{19.04}\stdev{0.53}             & \textbf{17.59}\stdev{0.49}             & \textbf{18.24}\stdev{0.49}            & \textbf{9.68}\stdev{0.37}             & \textbf{10.98}\stdev{0.41}            \\\hline
14K-\textit{light}                 & 10.31\stdev{0.38}           & 10.92\stdev{0.43}            & 9.83\stdev{0.33}            & 10.52\stdev{0.42}           & 4.96\stdev{0.31}            & 5.79\stdev{0.33}            \\
14K-\textit{large}                 & \textbf{18.93}\stdev{0.40}             & \textbf{18.97}\stdev{0.47}            & \textbf{17.22}\stdev{0.41}             & \textbf{18.12}\stdev{0.42}            & 9.03\stdev{0.35}             & 10.11\stdev{0.39}            \\
14K-\textit{xlarge}                & 18.14\stdev{0.42}    & 18.35\stdev{0.48}   & 15.90\stdev{0.39}    & 17.19\stdev{0.43}   & 5.46\stdev{0.29}    & 6.59\stdev{0.35}   \\\midrule
XLSR-53-\textit{large}~\citep{lebenchmark2}                 & 7.81\stdev{0.33}             & 6.75\stdev{0.29}             & 0.49\stdev{0.13}             & 0.52\stdev{0.08}             & 0.43\stdev{0.07}              & 0.36\stdev{0.05}             \\
XLSR-R-\textit{xlarge}                & 13.80\stdev{0.37}             & 13.88\stdev{0.38}            & 11.45\stdev{0.37}             & 12.56\stdev{0.40}            & 1.59\stdev{0.29}             & 1.77\stdev{0.31}     \\\bottomrule\\      
\end{tabular}
\label{ast:tab:extractionmtedx}
\end{table*}

\noindent\textbf{Hybrid results analysis and discussion.}  
Results are presented in Table~\ref{ast:tab:extractionmtedx} and analyzed with the following aspects: 
\begin{itemize}
    \item \textbf{Monolingual versus multilingual}. Comparing SSL models of the same size (large models), training on monolingual data (\nameproject models) seems to be beneficial in comparison with training on multilingual data (XLSR-53 and XLS-R models). From $3K$ hours of French data, all \nameproject large models outperform both XLSR-53 and XLS-R models. 
    \item \textbf{Pre-training data}. Concerning the amount of monolingual data, we observe that with the same model size (base or large), SSL models tend to improve when the amount of pre-training data increases, except for the 14K-large model whose performance is on par with that of the 7K-large model. We suspect that adding too much read speech data ($6,600h$) might lead to stagnation in terms of BLEU scores when jumping from $7K$ to $14K$ hours of training data on the mTEDx domain. 
    \item \textbf{Model size}. Table~\ref{ast:tab:extractionmtedx} illustrates that with the same amount of pre-training data, larger models tend to be better than smaller ones for producing speech features. Surprisingly, xlarge models underperform large models, observed with both \nameproject 14K and XLS-R. 
    Lastly, we observe that the 14K-light model significantly underperforms its base and large counterparts, hinting that it insufficiently represents speech signals due to its limited capacity.
\end{itemize}

\noindent\textbf{End-to-End results analysis and discussions.} 
Table~\ref{ast:tab:e2emtedx} and \ref{ast:tab:e2ecovost} present BLEU scores for the mTEDx and CoVoST datasets respectively. 
\begin{itemize}
    \item \textbf{Monolingual versus multilingual.} Overall, we notice that the importance of the backbone model (monolingual or multilingual) is less important in end-to-end mode compared to hybrid mode (the XLS-R model is not too far behind the best-performing monolingual \nameproject model). 
Nevertheless, between the two best-performing models for both CoVoST and mTEDx, \nameproject 14K outperforms XLS-R in all settings.
It should however be highlighted that XLS-R is a model covering 128 languages, with the potential to reach similarly good results for at least a small portion of its covered languages. 

\item  \textbf{Pre-training data.} Focusing on monolingual models, and looking at results for large architectures only, we do not see any hints of saturation: \nameproject models pre-trained with more speech data tend to provide better initializations for the AST task~(Tables~\ref{ast:tab:e2emtedx} and \ref{ast:tab:e2ecovost}). Looking at base architectures, the exception for this seems to be the 2.7K-base model, which performs on par with 3K and 7K base and large models. This model differs from 3K by not including spontaneous speech. We hypothesize this pre-training setting could provide a better initialization for mTEDx and CoVoST datasets, which are made of prepared and read speech respectively. 

\begin{table*}[]
\caption{AST end-to-end BLEU results (higher is better) for the mTEDx dataset. The best results are in \textbf{bold}. {\color{gray}Gray numbers} denote the standard deviation computed using bootstrap re-sampling~\citep{koehn2004statistical}.}
\centering
\begin{tabular}{lcc|cc|cc}\toprule
\multicolumn{1}{l}{}     & \multicolumn{2}{c}{\textbf{fr-en}} & \multicolumn{2}{c}{\textbf{fr-es}} & \multicolumn{2}{c}{\textbf{fr-pt}} \\
\multicolumn{1}{c}{\textbf{Representation}}  & \textbf{Dev}   & \textbf{Test}   & \textbf{Dev}   & \textbf{Test}   & \textbf{Dev}   & \textbf{Test}   \\\midrule
1K-\textit{base} & 15.2\stdev{0.48}  & 14.0\stdev{0.54}   & 13.0\stdev{0.42}  & 13.2\stdev{0.40}   & 8.2\stdev{0.33}   & 8.6\stdev{0.34}  \\
1K-\textit{large} & 16.7\stdev{0.49}  & 16.6\stdev{0.46} & 15.3\stdev{0.46} & 16.1\stdev{0.45} & 9.4\stdev{0.33} & 10.7\stdev{0.38} \\\hline
2.7K-\textit{base} & 18.9\stdev{0.52} & 18.7\stdev{0.52} & 17.9\stdev{0.50} & 17.8\stdev{0.49} & 11.7\stdev{0.39} & 12.3\stdev{0.40} \\\hline
3K-\textit{base} & 17.9\stdev{0.48} & 17.9\stdev{0.51} & 16.8\stdev{0.49} & 17.1\stdev{0.46}  & 11.3\stdev{0.42} & 12.4\stdev{0.42} \\
3K-\textit{large} & 17.6\stdev{0.51} & 16.9\stdev{0.47}  & 15.1\stdev{0.45} & 15.6\stdev{0.46} & 8.6\stdev{0.34} & 9.7\stdev{0.37} \\\hline
7K-\textit{base} & 18.8\stdev{0.51}  & 18.2\stdev{0.50} & 18.4\stdev{0.52} & 18.2\stdev{0.68}  & 12.6\stdev{0.41} & 13.4\stdev{0.44} \\
7K-\textit{large} & 20.1\stdev{0.52}  & 19.0\stdev{0.57}  & 17.4\stdev{0.52} & 18.8\stdev{0.49}  & 10.7\stdev{0.37} & 12.0\stdev{0.41} \\\hline
14K-\textit{light} & 6.5\stdev{0.27} & 5.9\stdev{0.28} & 5.7\stdev{0.27} & 5.7\stdev{0.26} & 3.0\stdev{0.21}  & 2.9\stdev{0.17} \\
14K-\textit{large} & 23.6\stdev{0.59} & 23.1\stdev{0.55} & 23.3\stdev{0.58}  & 24.2\stdev{0.62} & 18.7\stdev{0.54} & 21.8\stdev{0.58} \\
14K-\textit{xlarge}  & \textbf{25.1}\stdev{0.59}  & \textbf{24.4}\stdev{0.60} & \textbf{23.7}\stdev{0.56}  & \textbf{25.5}\stdev{0.59}  & \textbf{20.7}\stdev{0.58} & \textbf{23.7}\stdev{0.62} \\\midrule
XLSR-53-\textit{large} & 15.6\stdev{0.49}  & 12.5\stdev{0.47}  & 15.6\stdev{0.45}  & 15.8\stdev{0.44} & 8.4\stdev{0.31}  & 9.1\stdev{0.36} \\
XLS-R-\textit{xlarge}     & 23.4\stdev{0.58}  & 22.7\stdev{0.55} & \textbf{23.3}\stdev{0.61} & \textbf{25.0}\stdev{0.60}  & 19.3\stdev{0.54}  & 21.3\stdev{0.56}   \\\bottomrule\\
\end{tabular}
\label{ast:tab:e2emtedx}
\end{table*}

\begin{table}
\caption{ST end-to-end BLEU results (higher is better) for the CoVoST dataset. The best results are in \textbf{bold}. {\color{gray}Gray numbers} denote the standard deviation computed using bootstrap re-sampling~\citep{koehn2004statistical}.}
\centering
\begin{tabular}{llll}\toprule
\textbf{Representation}        & \multicolumn{1}{c}{\textbf{Dev}} & \multicolumn{1}{c}{\textbf{Test}} \\\midrule
1K-\textit{base}  & 28.5\stdev{0.21}   & 27.9\stdev{0.20}  \\
1K-\textit{large} & 30.1\stdev{0.21}   & 30.0\stdev{0.21}  \\\hline
2.7K-\textit{base} & 30.8\stdev{0.21}   & 30.2\stdev{0.21}  \\\hline
3K-\textit{base}  & 29.8\stdev{0.21}   & 29.4\stdev{0.21}  \\
3K-\textit{large} & 29.4\stdev{0.21}   & 29.0\stdev{0.21}  \\\hline
7K-\textit{base}  & 30.1\stdev{0.21}   & 29.7\stdev{0.20}  \\
7K-\textit{large} & 32.7\stdev{0.22}   & 32.5\stdev{0.21}  \\\hline
14K-\textit{light} & 20.5\stdev{0.18}       & 20.0\stdev{0.18}      \\
14K-\textit{large}  & 32.1\stdev{0.21}       & 31.7\stdev{0.21}      \\
14K-\textit{xlarge} & \textbf{33.9}\stdev{0.22}       & \textbf{33.7}\stdev{0.21}      \\\midrule
XLSR-53-\textit{large} & 30.4\stdev{0.21}       & 29.6\stdev{0.20}      \\
XLS-R-\textit{xlarge} & 32.9\stdev{0.21}       & 32.5\stdev{0.21}      \\\bottomrule\\
\end{tabular}
\label{ast:tab:e2ecovost}
\end{table}


\item \textbf{Model size.}
\textcolor{rebuttal}{Looking at mTEDx~(Table~\ref{ast:tab:e2emtedx}) and CoVoST results~(Table~\ref{ast:tab:e2ecovost}), we reach the overall conclusion that using larger pre-trained models as initialization for end-to-end AST tend to result in better AST performance. The only exceptions to these findings seem to be the 3K-large model applied to the mTEDx dataset, which results in AST models that are consistently worse than using 3K-base. Moreover, we also observe that for the mTEDx fr-pt setting, the 7K-large model is inferior to its base version. We believe these findings could hint that these two models (3K and 7K-large) are providing a suboptimal initialization for the AST task, which is more apparent in the mTEDx experiments, as in this setting we have considerably less trainable fine-tuning examples compared to CoVoST, and in particular for the languages where we observe the largest gaps (es and pt). 
}
Finally, it seems to always be beneficial for end-to-end AST fine-tuning to have a xlarge wav2vec~2.0 model, compared to large, but this marginal difference in performance adds a considerable overhead in the number of trainable parameters~(647.1M extra trainable parameters). Lastly, we observe that the 14K-light model is a poor initialization choice for end-to-end AST. We believe this highlights how the capacity of the model is related to the encoding of high-abstraction level speech features: smaller Transformer stacks results in poor speech features~(Table~\ref{ast:tab:extractionmtedx}) and encoders~(Tables~\ref{ast:tab:e2emtedx} and \ref{ast:tab:e2ecovost}). Indeed, Pasad et al.~\citep{pasad2021layer,pasad2023comparative} argue that the wa2vec~2.0 pretext task forces a drop in abstraction-level at the last layers. Due to this, the middle of the Transformer stack is where most of high-level (phonemic and word-level) information is encoded.

\end{itemize}

\subsection{Automatic Emotion Recognition}
\label{subsec:er}

%
%
%

Recent psychological studies suggest that emotion is needed and used in every aspect of our lives, from filtering the sensory information, our perception of an event, reasoning, and thus the decisions we make \citep{moors2013role, brosch2013impact}. The automatic recognition of human emotions from audio recordings is therefore a technology that can influence many areas such as education, healthcare, and entertainment. Although much progress has been made in emotion recognition in recent years, there are still challenges including different emotional expressions by different speakers, or different microphones, making AER not quite ready for everyday use \citep{alisamir2022multicorpus}. SSL models being trained on large amounts of data, have been shown to be exceptionally good in addressing generalisation issues \citep{lebenchmark2}. \nameprojectnew further evaluates such methods for French speech, with new trained SSL models. \\

\noindent\textbf{Downstream datasets.}
Following \nameproject, we used the RECOLA \citep{ringeval2013introducing} and the AlloSat \citep{macary2020allosat}, which contain continuous conversations, and the THERADIA corpus, which contains utterance-based conversations. Both the RECOLA and Allosat datasets contain spontaneous French speech. The RECOLA recordings are emotionally induced conversations recorded in a laboratory environment, whereas the AlloSat recordings are telephonic conversations. The annotations for both datasets are time-continuous dimensional. For Allosat, a frustration-to-satisfaction dimension is used with a sampling rate of 4\,Hz. And for RECOLA, the emotion dimensions are based on arousal (from passive to active), and valence (from negative to positive), sampled at 25\,Hz rate. Moreover, the AlloSat dataset contains a total of 29,704 utterances (21\,h), divided into 20,785 utterances (15\,h) as training set, 4272 utterances (3\,h) for development and 4643 utterances (3\,h) as test partition. On the other hand, the RECOLA dataset is much smaller, with 9 files of 5 minutes each for the training, development, and test sets. Since the continuous conversations used in RECOLA and AlloSat datasets differ from utterance-based training of the used SSL models, we also used the THERADIA corpus, which contains segments divided by utterances divided by pauses for breath. 

The THERADIA corpus contains 61 senior participants, nine of whom had Mild Cognitive Impairments (MCIs). The participants performed digital cognitively stimulating exercises while interacting with a virtual assistant in a natural way. The THERADIA corpus contains emotion labels that are annotated based on the perceived intensity of the label on a scale from zero (not existent), to 100. We report results on the prediction of the ten most common core set labels in the THERADIA corpus: relaxed, interested, frustrated, confident, satisfied, happy, annoyed, surprised, desperate, and anxious. The THERADIA corpus contains 2,735 utterances (6\,h) in total which are divided into 1,110 utterances as training partition, 851 utterances for validation, and 774 utterances for testing. \\


%
%
%
%

\noindent\textbf{Downstream models and hyperparameters.}
The experiments are conducted using a one-to-one sequence model with either SSL representations or Mel Filter Bank features. The experiments for the RECOLA and AlloSat datasets consist of time-linear sequence-to-sequence prediction of continuous dimensions of emotion. A GRU with one layer and 32 hidden units was trained with CCC as the loss function, similar to \citep{lebenchmark2}. Since both the AlloSat and RECOLA dataset contains continuous long audio files, we were not able to fine-tune them. On the other hand, for the experiments for the THERADIA corpus, we used one linear layer trained with mean squared error as the loss function. We also tried using CCC as the loss function and GRU as the emotion prediction model for THERADIA, but did not find any significant improvement in the results. Also, since THERADIA corpus contains separated utterances, we also investigated the effect of fine-tuning the SSL models for emotion recognition.

Furthermore, for all the experiments, the training was done with the Adam optimizer with a learning rate of 0.001 for frozen models, and 0.0001 for fine-tuning, and trained for 200 epochs with early stopping. It should be noted that that the sampling rate of the dimensional annotations differs from the Mel features, which are sampled at a rate of 100\,Hz, and the wav2vec representations, which are sampled at a rate of 50\,Hz. Thus, during training for RECOLA and AlloSat datasets, the targets are resampled to match the sampling rate of the representations with linear interpolation to keep the graph active for backpropagation, while during testing, the outputs of the model are mapped to the sampling rate of the targets to keep the targets untouched for testing. On the other hand, for the THERADIA corpus, the outputs of the model are averaged over the sequence, because each emotion label is defined per sequence and not continuously over the sequence.\\


\noindent\textbf{LeBenchmark 1.0.} \textcolor{rebuttal}{In the previous version of the AER benchmark, we have investigated the use of SSL models only for acoustic feature extraction using the AlloSat and RECOLA datasets. It should also be noted that the difference between the results of the previous work and this paper is due to the change of the architecture and framework used for training the SSL models. This paper also presents additional results for the THERADIA corpus for both the use of SSL models for feature extraction (referred to as ``frozen'') and end-to-end fine-tuning (referred to as ``fine-tuned'').} \\

\begin{table}[]
\caption{The AER results are expressed in terms of concordance correlation coefficient (CCC, higher is better). \textcolor{rebuttal}{The left table describes the continuous prediction of frustration-satisfaction, arousal and valence dimensions for the AlloSat and RECOLA corpora (for concatenated outputs), with frozen representations~(feature extraction setting). The results on the right table describe the average~({\color{gray}gray numbers} describe standard deviation) emotion prediction across the core set emotion labels of the THERADIA corpus in the frozen~(feature extraction) and fine-tuned~(end-to-end).} } 
\centering
\begin{tabular}{cc}
\scalebox{0.95}{
\begin{minipage}{.5\linewidth}
    \begin{tabular}{lccc}
        \multicolumn{4}{c}{Corpora: AlloSat \& RECOLA} \\
    \multicolumn{4}{c}{Metric: Concordance Correlation Coefficient (CCC) $\uparrow$} \\
        \toprule
        \textbf{Representation} & \textbf{Satisfaction} & \textbf{Arousal} & \textbf{Valence} \\
        \hline
        Mel Filter Bank & .413 & .313 & .258 \\
        \hline
        1K-\textit{base} & .487 & .427 & .055 \\
        1K-\textit{large} & .021 & .018 & .001 \\
        \hline
        2.7K-\textit{base} & .596 & .629 & .455 \\
        \hline
        3K-\textit{base} & .602 & .358 & .007 \\
        3K-\textit{large} & .040 & .097 & .000 \\
        \hline
        7K-\textit{base} & .470 & .335 & .116 \\
        7K-\textit{large} & .050 & .009 & .037 \\
        \hline
        14K-\textit{light} & .518 & .614 & .348 \\
        14K-\textit{large} & .462 & \textbf{.664} & \textbf{.466} \\
        14K-\textit{xlarge} & \textbf{.657} & .649 & .437 \\
        \midrule

        \textcolor{rebuttal}{XLSR-53-\textit{large}} & .264 & .149 & .146 \\
        XLS-R-x\textit{large} & .415 & .311 & .229 \\
        \bottomrule\\
    \end{tabular}
\end{minipage} 
}
\scalebox{0.95}{
\begin{minipage}{.5\linewidth}
    \vspace{-0.4cm}
    \begin{tabular}{lcc}
        \multicolumn{3}{c}{} \\
        \multicolumn{3}{c}{Corpus: THERADIA} \\
        \multicolumn{3}{c}{Metric: Concordance Correlation Coefficient (CCC) $\uparrow$} \\
        \toprule
        \textbf{Representation} & \textbf{Frozen} & \textbf{Fine-tuned} \\
        \hline
        Mel Filter Bank & .075\stdev{.120} & - \\
        \hline
        1K-\textit{base} & .151\stdev{.103} & .246\stdev{.161} \\
        1K-\textit{large} & .001\stdev{.002} & \textbf{.319\stdev{.172}} \\
        \hline
        2.7K-\textit{base} & .083\stdev{.094} & .013\stdev{.015} \\
        \hline
        3K-\textit{base} & .061\stdev{.069} & .019\stdev{.016} \\
        3K-\textit{large} & .002\stdev{.004} & .224\stdev{.151} \\
        \hline
        7K-\textit{base} & .106\stdev{.082} & .000\stdev{.000} \\
        7K-\textit{large} & .000\stdev{.002} & .230\stdev{.143} \\
        \hline
        14K-\textit{light} & \textbf{.241\stdev{.133}} & .283\stdev{.129} \\
        14K-\textit{large} & .190\stdev{.127} & .229\stdev{.151} \\
        14K-\textit{xlarge} & .237\stdev{.131} & .226\stdev{.145} \\
        \midrule
        \textcolor{rebuttal}{XLSR-53-\textit{large}} & .004\stdev{.006} & .232\stdev{.144} \\
        XLS-R-\textit{xlarge} & .072\stdev{.059} & .205\stdev{.132} \\
        \bottomrule\\
    \end{tabular}
\end{minipage}
}
\end{tabular}
\label{tab:AER-Dim}
\end{table}

\noindent\textbf{Results analysis and discussion.} 
\textcolor{rebuttal}{The results are shown in Table~\ref{tab:AER-Dim}. Overall, the results across different 1K, 3K, and 7K models seem to vary a lot, while the results for 14K models seem to be consistently good for different emotion recognition tasks. For example, it can be observed that the 1K-large, 3K-large and 7K-large models perform poorly when frozen. This result is not limited to the AER tasks in this section, but also extends to Section~\ref{subsec:sa}~(see WER in Table \ref{sa:tab:result}). This may indicate that these models were trained in such a way that does not allow them to 
generalise to the AER task in an off-the-shelf setting. 
Further investigation is needed to determine the reason for this. 
}

\textcolor{rebuttal}{Regarding 14K models, results tend to vary less across different experiments, and are consistently better than traditional features. This suggests that the pre-training stage is not suffering of saturation, and that adding more data at that stage results in a more robust off-the-shelf model, which can achieve consistently better results for AER than traditional features. We can also observe that all the 14K models trained on French speech, when frozen, perform better than the multilingual SSL models (XLSR-53-large and XLSR-R-xlarge). 
}


\textcolor{rebuttal}{Focusing on AlloSat and RECOLA results, we notice that for the prediction of emotion dimensions, the 14K-large and 14K-xlarge models achieved the best scores. While these two models achieved similar results for the prediction of arousal and valence dimensions of emotion on the RECOLA corpus, their results are not similar for the frustration-satisfaction dimension of the AlloSat corpus. This may be because RECOLA contains clean speech, and thus the smaller parameter size of 14K-large, compared to 14K-xlarge, may be sufficient to extract useful features for continuous emotion recognition on this dataset. However, the better results of the 14K-xlarge model compared to 14K-large for the AlloSat corpus, might suggest that for the real-life telephonic conversations, the use of a larger model is beneficial. 
}

\textcolor{rebuttal}{
Similarly, the prediction of emotion labels for the THERADIA corpus with the frozen SSL models shows that the 14K models perform best. However, when fine-tuning the wav2vec 2.0 models, we observe that the base models (except for 1K-base) perform worse than their frozen counter parts. On the other hand, for the large models (except for 14K-xlarge), the performance is improved when fine-tuning the SSL models. The better performance of the fine-tuned wav2vec 2.0 models is consistent with the literature. Indeed, fine-tuning specializes the representations, which results in better AER performance for a particular data distribution. However, we may lose the ability to generalize to other data distributions~\citep{alisamir2022multi}. Moreover, the fact that 1K-large model performs best when fine-tuned, may suggest that pre-training the wav2vec 2.0 models with more data or more parameters does not necessarily have a one-to-one correlation with better prediction performance of emotion labels. This, in turn, may mean that predicting THERADIA's emotion labels without regard to generalisation does not necessarily require complex architectures or large amounts of data to train and then fine-tune a base model. 
}
\subsection{Syntactic Analysis}
\label{subsec:sa}

%
%
%
The syntactic analysis task (also known as parsing) is a staple task in \textcolor{rebuttal}{NLP}, 
and historically one of the first.
Syntactic parsing consists of assigning a syntactic structure to a given sentence.
Thus, it is a structured prediction task. 
Corpora annotated with syntactic trees are key to data-driven linguistic studies, syntactic trees may also provide useful features for downstream tasks. We focus on evaluating the 
\textcolor{rebuttal}{pre-trained SSL models} trained on the task of joint \textcolor{rebuttal}{ASR} 
and syntactic analysis. The traditional technique to obtain the syntactic structure from speech would be to use a pipeline approach.
Using an ASR model to get the transcription and then using a pre-trained model such as BERT to predict the syntactic structure.
However, this method removes important features contained in the signal for the syntactic predictions, such as the prosody. 
Moreover, it has been shown that \textcolor{rebuttal}{end-to-end} 
speech parsing models perform better, despite having much fewer parameters than pipeline-based parsers \citep{pupier22_interspeech}. \\

%
%
%
%

\noindent\textbf{Downstream datasets.} 
We use the CEFC-ORFEO \citep{benzitoun2016projet} dataset. This corpus is a collection of multiple subcorpora\citep{cfpp2000,11403/clapi/v1,11403/tcof/v2.1, OFROM, Fleuron, FON, Coralrom, delic:halshs-01388193} annotated in syntactic dependencies in the \textit{conll} format. 
This corpus contains a wide variety of speech situations,
such as store owner/customer interactions in a cheese shop, or informal conversations between friends.
We removed the TCOF sub-corpus from the dataset, as it was included in the pretraining data for the \nameprojectnew models.
The Orféo treebank contains both human-annotated syntactic trees and automatically annotated
syntactic trees \citep{Orfeo-annotation-auto}, henceforth referred to as silver data. 
Partitions are built such that the dev and test set only contain gold trees, i.e.\ all silver trees are in the training set. \\

%
%
%
%

\begin{table}[t]
\caption{End-to-end results for the SA task in terms of part of speech tagging f1-score, unlabeled attachment score (UAS), and labeled attachment score (LAS) metrics (higher is better). Results are correlated to the speech recognition results expressed in CER and WER. Syntactic analysis training started only for model with WER below the threshold of 50. }
\centering

\begin{tabular}{lcccccc}
\toprule
\textbf{Representation}     & \textbf{Frozen} & \textbf{WER \%}   & \textbf{CER \%}   & \textbf{POS}  & \textbf{UAS}   & \textbf{LAS} \\
\midrule
1K-\textit{base}    &No& 51.26 & 31.32 & —     & —     & —           \\
           &Yes& 80.35    & 48.67    & —     & —     & —          \\
\hline
1K-\textit{large}   &No& 45.99 & 28.60 & 65.18 & 59.98 & 53.48     \\
           &Yes& 99.82    & 93.80    & —     & —     & —        \\
\hline
\textcolor{rebuttal}{2.7K-\textit{base}}   &No& 100 & 88.40 & —  & —  & —      \\
           &Yes& 93.97    & 57.95   & —     & —     & —        \\
\hline

3K-\textit{base}    &No& 100    & 88.40     & —     & —     & —            \\
           &Yes& 81.54     & 48.16     & —     & —     & —         \\
\hline
3K-\textit{large}   &No& 44.81 & 27.55 & 65.81 & 61.54 & 55.11     \\
           &Yes& 99.94    & 88.33    & —     & —     & —     \\
\hline
7K-\textit{base}    &No& 100    & 88.40     & —     & —     & —          \\ 
           &Yes& 70.70     & 39.97     & —     & —     & —      \\
\hline
7K-\textit{large}   &\textbf{No}& \textbf{42.39} & \textbf{26.59} & \textbf{67.15} & \textbf{63.34} & \textbf{56.94}    \\
           &Yes& 99.81     & 78.59     & —     & —     & —     \\
\hline
14K-\textit{light}  &No&71.08 &40.51 &— &— &—  \\
                    &Yes&76.56    & 46.28     & — & — & —  \\
\hline
14K-\textit{large}  &No& 43.04 & 27.12 & 66.71 & 62.72 & 56.45       \\
                    &Yes&  52.16 & 30.51     & —     & —     & —           \\
\hline
14K-\textit{xlarge} &No&  44.82  & 28.00     &   65.09    &    61.17   & 54.61       \\ 
                    &Yes& 45.43  & 26.51 & 66.60      &  60.63     &  53.94  \\\hline
XLSR-53-\textit{large} & No & 44.57 & 26.77 & 66.40 & 61.20 & 54.69\\
& Yes & 97.51 & 77.81 & — & — & —  \\ 
\hline
\textcolor{rebuttal}{XLS-R-\textit{xlarge}} &No&  54.88  & 32.50     &   —    &    —   & —        \\ 
                    &Yes& 99.16  & 76.08 & —       &  —      &  —   \\ \bottomrule
\hline
\end{tabular}
\label{sa:tab:result}
\end{table}

\noindent\textbf{Downstream models and hyperparameters.} 
%
%
%
%
The downstream model is wav2tree \citep{pupier22_interspeech}.
This model is composed of an encoder module, an ASR module, and a syntax module. It performs joint syntactic analysis and automatic speech recognition in a multitasking manner. 
The encoder is composed of three fully connected layers of size 1024 in the fine-tuning setting.
In the frozen setting, the encoder is a two-layer bi-LSTM with a hidden size 1024 and a dropout of 0.4 between the two layers. In wav2tree, the speech recognition module has two purposes, the first one is the normal speech recognition task, acquiring the transcriptions.
The second one is the segmentation of the representation learned by wa2vec~2.0. The CTC scheme labels each representation with either a letter, a blank, or a whitespace. The wav2tree model uses the  whitespace information to segment the representation. The syntax module is composed of two elements. The first one uses the segmentation from the ASR module to create word-level representations from the encoder representation via a 2-layer LSTM with a hidden size of 500. These word-level representations are then used by a two-layer bi-LSTM with a hidden size of 800 and then are classified into three tasks in parallel with a simple linear layer. The first one predicts the part of speech (POS) of the word, the second one predicts the head of the current word (UAS) and the last one predicts the syntactic function (e.g. subject, dependent, etc), i.e.\  the relationship between the head and the dependent. Each model is trained with a batch size of 8, except for the fine-tuning of the xlarge which is trained with a batch size of 2 and a gradient accumulation of 4, in order to maintain the comparability of results. The ASR module is trained with a CTC loss and the classification tasks are trained with a negative likelihood loss. The optimizer is AdaDelta \citep{DBLP:journals/corr/abs-1212-5701} with a learning rate of 1 and $\rho$ of 0.95. Each model is optimized on the word error rate and once it decreases below a threshold of 50, the training activates the syntax learning and is then optimized on the LAS metrics. \\

\noindent\textbf{Parsing evaluation metrics.}
The three metrics used for the parsing task are the f-measure for part of speech tagging labeled POS, the unlabeled attachment score UAS which is the accuracy of the syntactic link between the word, i.e. is the word correctly attached to its head.
The last one is the labeled attachment score LAS extending the UAS metrics by also taking into account the nature of the link (root, subject, etc).
\textcolor{rebuttal}{Since the ASR transcript differ from the gold transcript, an alignment between the two must be done in order to compute these different metrics.
We use the same methods as \citep{pupier22_interspeech}.
By using the WER alignment computed during the evaluation of the ASR module, we can realign the two transcript and compute these metrics.}\\

\noindent\textbf{Results analysis and discussion.} 
All results for the syntactic analysis task are depicted in Table~\ref{sa:tab:result}.
The parsing result is heavily correlated to the \textcolor{rebuttal}{ASR} 
metrics. 
This is an expected behavior, since a correct tree cannot be produced if some words are missing.
The WER and CER clearly reflect the \textcolor{rebuttal}{challenge} 
of the target dataset, 
\textcolor{rebuttal}{as with a similar architecture, a score of 10 \% of WER is obtained on CommonVoice~6.1 \citep{ardila-etal-2020-common}.} 
The difficulty of the dataset implies that none of the base models can get good enough results to start learning the parsing task.
All the models also need to be fine-tuned on the dataset with the notable exception of the 14K-xlarge suggesting that the pre-trained representation of this model is general enough to fit more out-of-domain data like the CEFC-Orfeo dataset.

We observe that the quantity of data used to pre-train the model is important and seems to follow the classic machine learning paradigm that more data and scale are better.
However, the best model for this task is the 7K-large and not the 14k-large or xlarge. 
Our hypothesis is that this model is trained on a more balanced distribution of type of speech (read, prepared and spontaneous), thus being more suited to learning good representations for spontaneous speech. 
Another interesting fact is that the 14k-large outperforms the 14k-xlarge. 
This may be simply because the bigger model needs more data, whereas the smaller one is more easily tunable to the downstream dataset.
One of the most surprising results is the one on the 3K-base and 7k-base, where the models perform better without fine-tuning. We also compare to the multilingual XLSR-53 model.
The multilingual model has slightly worse performance compared to most of our models, but exhibits similar properties, such as the need to finetune it on the downstream corpus to reach good performance. 
\textcolor{rebuttal}{The smaller multilingual model achieves better performance than the bigger one, XLSR-R-xlarge. 
This may be a combination of the same phenomenon as the one observed on the 14k-large and 14k-xlarge and the difference of the two dataset used to train these model. 
XLS-R-xlarge is trained on 128 languages, whereas XLSR-53 on 53.
The added languages may negatively impact the performance on spontaneous speech since spontaneous speech corpora are less common (even unannotated), thus unbalancing the distribution between the different type of speech.
}

\subsection{Automatic Speaker Verification}
\label{subsec:asv}

Automatic Speaker Verification (ASV) refers to the task of verifying the identity claimed by a speaker from that person's voice~\citep{bimbot2004tutorial}. In ASV, deep neural networks have brought about significant advancements in voice representations, outperforming the previous state-of-the-art $i$-vector framework~\citep{dehak2010front}. One of these DNN approaches seeks to extract a high-level speaker representation, known as \textit{speaker embedding} directly from acoustics excerpts. To achieve this, DNN models are trained through an ASV task, where speech segments are classified into distinct speaker identities. Each layer of the DNN is trained to extract relevant information for discriminating between different speakers, and one of the hidden layers is used as the speaker embedding. One of the main advantages \textcolor{rebuttal}{of this approach} is that speaker embeddings produced by the DNN can generalize well to speakers beyond those present in the training set. The benefits of speaker embedding, in terms of speaker detection accuracy, have been demonstrated during the last evaluation campaigns: NIST SRE~\citep{villalba2018jhu,lee2019nec,rouvier2019} and VoxCeleb 2020~\citep{thienpondt2020idlab,brummerbut+,torgashov2020id}.

Recently, much progress has been achieved in ASV through the utilization of SSL models. In~\citep{chen2022large}, the authors proposed a novel approach: instead of exclusively using the representations from the final layer of the SSL model, they employ a weighted average of the representations from all hidden layers of the SSL model. This approach allows for harnessing speaker-related information embedded throughout the entire SSL model. \textcolor{rebuttal}{More recently, the authors in~\citep{vaessen2022fine} investigated methods to fine-tune a pre-trained wav2vec2 model specifically for ASV. They proposed a fine-tuning approach that introduces a CLS token at the beginning of the process for utterance-pair classification.} \\

\noindent\textbf{Downstream datasets.} For evaluation, we used the Fabiole dataset~\citep{ajili2016fabiole}. Fabiole is a french speaker verification dataset that has been collected for use to highlight the importance of the ``speaker factor" in forensic voice comparison. It contains 7,372 segments from 130 male native French speakers. Due to the absence of an established evaluation protocol, we took the initiative to create one ourselves. We removed all the segments with less than 2 seconds of voice and those that exceed 12 seconds. 
Then, we randomly selected 300,000 target pairs (i.e. enrollment and test segments that target same speakers) and 300,000 non-target pairs (i.e. enrollment and test segment that non-target same speaker). We trained the systems using ESTER-1~\citep{galliano2006corpus}, ESTER-2~\citep{galliano2009ester}, ETAPE~\citep{gravier2012etape} and REPERE~\citep{giraudel2012repere} training datasets (that correspond to 2.911 speakers and more than 250 hours of data). Voice Activity Detection (VAD) processing was not applied on the training datasets. Additionnaly, we applied data augmentation into the training process by incorporating noise from the MUSAN dataset and reverberation using the RIR dataset~\citep{snyder2015musan}. Equal Error Rate (EER) and the Detection Cost Function (DCF) are used as the performance criterion of ASV. EER is the threshold value such that the false acceptance rate and miss rate are equal. Whereas DCF is a weighted sum:

\begin{equation}
\begin{array}{l}
    C_{det} = C_{Miss} \times P_{Miss|Target} \times P_{Target} + 
    C_{FalseAlarm} \times P_{FalseAlarm|NonTarget} \times P_{NonTarget},
\end{array}
\end{equation}

with the prior probabilities $P_{Target}$ and $P_{NonTarget} = 1 - P_{Target}$ of target and impostor speakers, respectively. The relative costs of detection errors in this function are the costs of miss $C_{Miss}$ and false alarm errors $C_{FalseAlarm}$. These parameters were set as follows: $P_{Target} = 0.01$ (or $P_{Target} = 0.001$), $C_{Miss} = 1$ and $C_{FalseAlarm} = 1$. \\

\noindent\textbf{Downstream models and hyperparameters.} We use the ECAPA-TDNN classifier~\citep{desplanques2020ecapa}. This classifier, when combined with SSL models, has demonstrated impressive performance~\citep{thienpondt2021idlab} in ASV. Our ECAPA-TDNN has the following parameters: the number of SE-Res2Net Blocks is set to 3 with dilation values 2, 3 and 4, the number of filters in the convolutional frame layers is set to 512 and embedding layer size is set to 256. The training lasted for 8 epochs with the \textit{AdamW} optimizer. We trained all the models with Circle Margin Loss and set the margin to 0.35. During the training process, we randomly sampled 3s segments from each utterance to construct a training batch. We remind that the ECAPA-TDNN takes as input a weighted average of the representations from all hidden layers of the SSL model. \\

\begin{table}[!t]
\caption{Results for the downstream task of ASV. Performance are expressed in terms of Equal Error Rate (EER, lower is better) and Minimum of the Detection Cost Function (minDCF, lower is better).}
\begin{center}
    \begin{tabular}{lccc}
        \multicolumn{4}{c}{Corpora: Fabiole} \\
        \multicolumn{4}{c}{Metric: EER and minDCF $\downarrow$} \\
        \toprule
        \textbf{Representation} & \textbf{EER} & \textbf{minDCF$^{-10}$} & \textbf{minDCF$^{-100}$} \\
        \midrule
        1K-\textit{base} & \eer{8.272} & \dcf{0.5557} & \dcf{0.7218} \\
        1K-\textit{large} & \eer{6.747} & \dcf{0.5084} & \dcf{0.7053} \\
        \midrule
        3K-\textit{base} & \eer{4.822} & \dcf{0.3737} & \dcf{0.5672} \\
        3K-\textit{large} & \eer{5.06} & \dcf{0.3743} & \dcf{0.5213} \\
        \midrule
        7K-\textit{base} & \eer{4.728} & \dcf{0.3636} & \dcf{0.5379} \\
        7K-\textit{large} & \eer{5.228} & \dcf{0.3833} & \dcf{0.5754} \\
        \midrule
        14K-\textit{light} & \eer{7.389} & \dcf{0.5079} & \dcf{0.7111} \\
        14K-\textit{large} & \eer{3.535} & \dcf{0.2965} & \dcf{0.4801} \\
        14K-\textit{xlarge} & \textbf{\eer{2.9}} & \textbf{\dcf{0.2411}} & \textbf{\dcf{0.4164}} \\
        \midrule
        XLSR-53-\textit{large} & \eer{6.676} & \dcf{0.4923} & \dcf{0.6771} \\
        XLS-R-\textit{xlarge} & \eer{6.674} & \dcf{0.4568} & \dcf{0.6331} \\        
        \bottomrule\\
    \end{tabular}
\end{center}
\label{tab:asv_result}
\end{table}
\begin{figure}[!t]
\center
\includegraphics[height=250px]{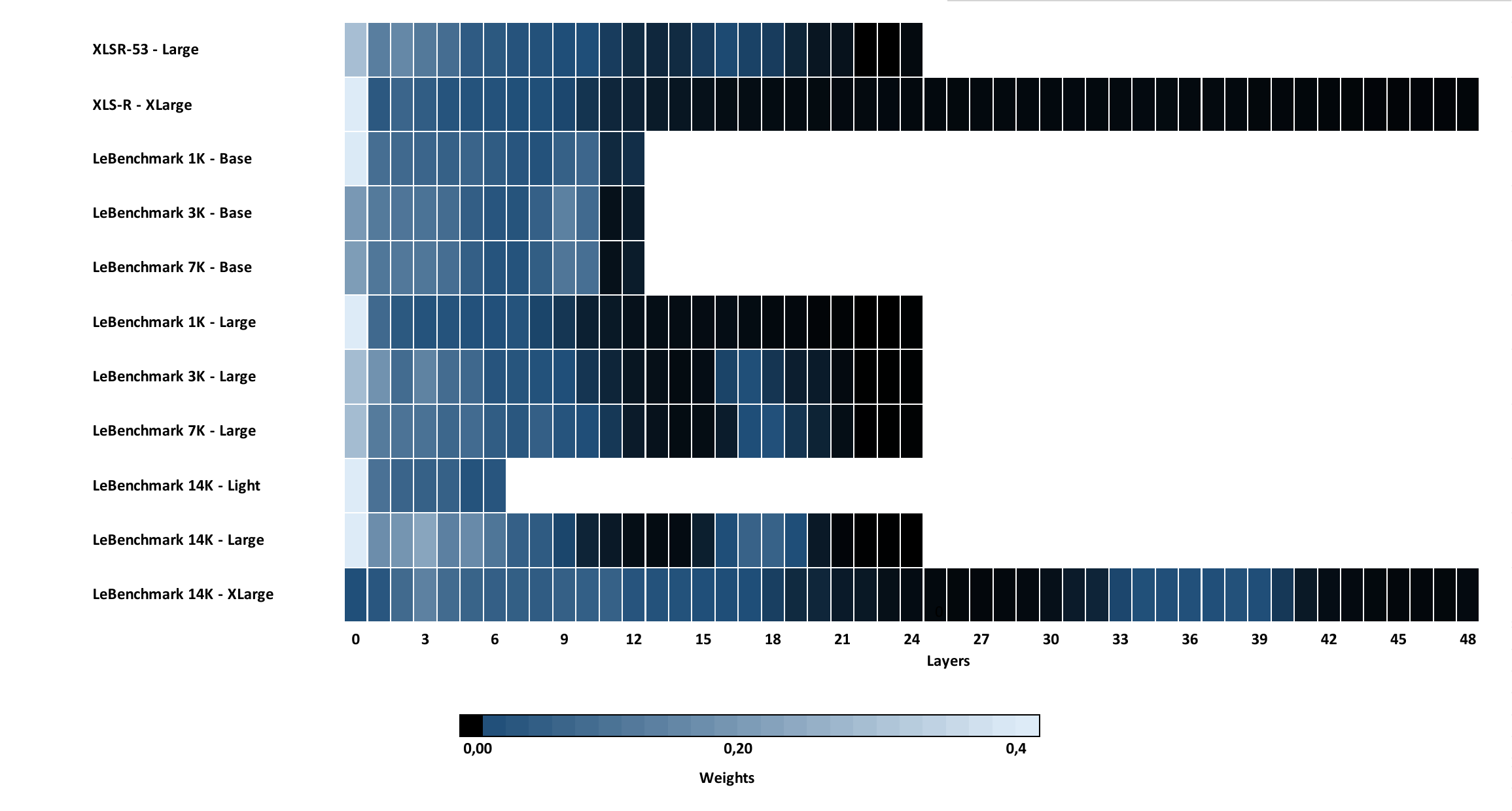}
\caption{The visualization of the normalized weight values in the proposed architecture. Each weight can be interpreted as the relevance of a given layer for the ASV task. Earlier layers are particularly relevant for ASV.}
\label{fig:heatmap_asv}
\end{figure}

\noindent\textbf{Results analysis and discussions.} All results for the ASV task are depicted in Table~\ref{tab:asv_result}. \textcolor{rebuttal}{XLSR-53-\textit{large} and XLS-R-\textit{xlarge} were} used as a baseline. The findings can be summarized as follows:

\begin{itemize}
    \item \textbf{Monolingual versus Multilingual.} We observed that except for 1K models, systems trained on monolingual models (i.e. \nameproject) achieved better performance than the multilingual model \textcolor{rebuttal}{(XLSR-53-\textit{large} and XLS-R-\textit{xlarge})}. The best monolingual model (\nameproject-14K-\textit{xlarge}) obtained 2.90\% of EER while the multilingual model (XLS-R-\textit{xlarge}) obtained 6.67\% EER.

    \item \textbf{Pre-training data.} Focusing on monolingual models, we observed a link between performance and the quantity of pre-training data. LeBenchmark models pre-trained with larger speech datasets tend to provide better performance. Indeed, the LeBenchmark model trained on 1,000 hours of data (LeBenchmark-1k-large) obtained 6.75\% of EER, while the LeBenchmark model trained on 14.000 hours of data (LeBenchmark-14k-xlarge) obtained 2.90\% of EER.

    \item \textbf{Model size.} \textcolor{rebuttal}{Always focusing on monolingual models, \textit{large} or \textit{xlarge} models obtain better performance than \textit{basic} models, except for 3K models and 7K models.} Figure~\ref{fig:heatmap_asv} shows the contribution from each layer of various SSL models. \textcolor{rebuttal}{We remind that LeBenchmark base models contain 12 layers, large models contain 24 layers and xlarge models contain 48 layers.} In general, we observe that speaker-related information is most pronounced in the first layers (lower layers) of the SSL model. Even if the speaker-related informations is most pronounced in the first layer, we notice that for LeBenchmark base models, all layers contribute to the construction of the representations. In contrast, for LeBenchmark large or xlarge models, the higher layers contribute less compared to the lower layers.

\end{itemize}

\textcolor{rebuttal2}{\subsection{Inference cost considerations}}
\label{subsec:inference}

\textcolor{rebuttal2}{Once an SSL model has been pre-trained and fine-tuned, it may be deployed in many scenarios with many different hardware environments. Reporting exhaustively the cost of inferring over our SSL models is intractable. A few preliminary works tried to approach this problem by estimating or analyzing the cost of inferring with pre-trained models. Still, it always remains tied to a specific use-case of deployment \citep{luccioni2023power}. This section provides a basic inference cost analysis based on CPU and GPU inference over the four different sizes of model provided by \nameprojectnew.\\}

\noindent\textcolor{rebuttal2}{\textbf{Task definition and measurements.} Among the SSL models of \nameprojectnew, the only factor affecting the inference cost is the size. Hence, the analysis is conducted by reporting different metrics obtained over the processing of one hundred sequences of length 15s by the following models: 14K-\textit{light}, 1K-\textit{base}, 14K-\textit{large} and 14K-\textit{xlarge}. The real-time factor (RTF) was obtained by taking the average time necessary to infer over 100 sentences divided by the length of one sequence. Hence, lower is better. The peak VRAM is also included. RTFs are reported both for CPU and GPU inference. Measurements are conducted on an isolated compute node with a single Nvidia RTX 3090 and 8 cores of an Intel Xeon 5218. Despite a higher number of cores being available, we fixed it to 8 to remain close to a relatively low-resource hardware setup. No optimization of the pre-trained model is performed. The checkpoint are simply loaded in SpeechBrain and used for inference.\\}

\noindent\textcolor{rebuttal2}{\textbf{Results analysis and discussion.} Table \ref{tab:inference_result} reports the obtained RTFs and the peak VRAM for every model. As one may expect, the bigger the model, the higher the RTF, and the slower the inference. This must be put into perspective with the obtained downstream performance of the previous sections. Indeed, the very small accuracy gains obtained with the 14K-\textit{xlarge} may not be sufficient to overcome the massive increase in inference cost, especially on CPU. For instance, 14K-\textit{large} is more than twice as fast as 14K-\textit{xlarge} and its downstream performances remain remarkable. As expected, GPU inference does not suffer from this distinction from the model size as the RTF remains extremely small in all scenarios. The peak VRAM, however, increases drastically as a GPU with under 8GB of VRAM would not be able to handle the 14K-\textit{xlarge} natively.} \\

\begin{table}[!t]
\caption{\textcolor{rebuttal2}{Real-time factor (RTF) and peak VRAM observed for the inference of 100 sentences of length 15s on a single RTX 3090 and 8 cores of an Intel Xeon CPU. Lower is better.}}
\begin{center}
    \begin{tabular}{lccc}
        \toprule
        \textbf{Model} & \textbf{CPU RTF $\downarrow$} & \textbf{GPU RTF $\downarrow$} & \textbf{Peak VRAM $\downarrow$} \\
        \midrule
        14K-\textit{light} & 0.028 & 0.001  & 1.2 GB  \\
        1K-\textit{base} & 0.038 & 0.001 & 2.2 GB  \\
        14K-\textit{large} & 0.096 & 0.002 & 5.1 GB  \\
        14K-\textit{xlarge} & 0.229 & 0.003 & 11.5 GB \\
        \bottomrule\\
    \end{tabular}
\end{center}
\label{tab:inference_result}
\end{table}

\subsection{Summary of results}
\label{subsec:discussion}

\textcolor{rebuttal}{
In the preceding sections, we presented multiple configurations along with their respective outcomes for the six downstream tasks introduced in \textit{LeBenchmark 2.0}. Unlike SUPERB, which primarily focused on evaluating SSL models with shallow probes, our approach aimed to assess our SSL models using diverse strategies applicable to downstream tasks. In this section, we summarize the key discoveries and insights derived from our exploration.}\\

\noindent\textbf{SSL as feature extractor (SLU, AST, AER, ASV). } \textcolor{rebuttal}{In this setting, downstream tasks exploited the representations produced by SSL models without adapting them during downstream training. However, for all mentioned tasks except ASV, this setup does not align with the SUPERB settings: representations are extracted from the penultimate layer instead of being weighted across the output generated by different layers. Overall, we find that in this setting, models trained with more hours of speech tend to yield the best results. For SLU, AST, AER, and ASV, the best SSL model for feature extraction was a 14K model, but the preferred feature dimension varied: SLU, AER and ASV achieved their best results with 14K-xlarge, while AST achieved them using 14K-large.}\\

\noindent\textbf{SSL as speech encoder (ASR, AST, AER, SA).} \textcolor{rebuttal}{In this setting, the SSL models were updated (fine-tuned) during downstream training, serving as speech encoders. The trend of results in this setting is not consistent. For ASR (CommonVoice and ETAPE datasets), we observe that 14K models are unable to outperform results obtained by models trained with less data: for both datasets, 3K and 7K-large models achieve the best results. Similarly, SA reached their best scores with the 7K-large model.
For the AST task (mTEDx and CoVoST), the best results are achieved by using the 14K-xlarge model. Surprisingly, for the AER task (THERADIA), the best-performing model was the 1K-large. We believe that this inconsistency in model ranking illustrates that probing rankings should not be used for selecting a model for end-to-end fine-tuning. Indeed, in \citet{zaiem2023benchmark}, the authors observed that model ranking quickly changes when downstream settings vary. We also believe that aspects such as the mismatch of pre-training and downstream datasets, the target task, the chosen architecture, and downstream optimization may all play key roles in the final performance of models.}\\

\noindent\textbf{SSL as feature extractor vs SSL as speech encoder~(AST, AER).} \textcolor{rebuttal}{In cases where both feature extraction and end-to-end fine-tuning approaches are employed for downstream tasks, we observe that feature extraction performance tends to lag behind end-to-end fine-tuning, as seen in AST and AER. This outcome is anticipated, given that the latter benefits from adapting the entire SSL model representation to the target data. However, this advantage comes at a higher training cost, also requiring more fine-tuning examples. Moreover, it appears that our multilingual baselines (XLSR-53-large and XLS-R-xlarge) particularly benefit from fine-tuning, as it may enable them to specialize.}\\


\noindent\textcolor{rebuttal}{\textbf{Overall simplified SSL model ranking (task-wise).}}
Table \ref{tab:summary2} summarizes the best models for each evaluated task and setup based on our experiments. Overall, the 7K and newly introduced 14K models 
outperform others across the 
benchmark. In all instances, larger models achieve superior performance. 
\noindent\textcolor{rebuttal}{The smaller models, base and light,} exhibit relatively lower performance, yet still within an acceptable range. In all scenarios, \nameprojectnew models demonstrate a higher level of performance than the \noindent\textcolor{rebuttal}{multilingual baselines~(XLSR-53-base and XLS-R-xlarge). This highlights that language-specific SSL models are still a better choice for application in downstream tasks, compared to general purpose multilingual models.}

%

\begin{table}[!t]
\caption{Summary of the best SSL models for each downstream task.}
\begin{center}
\begin{tabular}{ccc}
        \begin{tabular}{ccc}
            \toprule
            \textbf{Task} & \textbf{Best Feature Extractor} & \textbf{Best Speech Encoder} \\
            \hline
            
            ASR & - & \{3K-7K\}-\textit{large}  \\
            SLU & 14K-\textit{xlarge} & - \\
            AST & 7K-\textit{large} & 14K-\textit{xlarge}  \\
            AER  & 14K-\{\textit{light}, \textit{large}, \textit{xlarge}\}& 1K-\textit{large}  \\
            SA & -  & 7K-\textit{large}\\
            ASV & 14K-\textit{xlarge} & -  \\
            \bottomrule\\
        \end{tabular}    
\end{tabular}
\end{center}
\label{tab:summary2}
\end{table}

\section{Carbon Footprint}
\label{sec:footprint}

This section gives an overview of the carbon footprint of the SSL pre-training. The fine-tuning footprint is omitted as it was performed in many different and heterogeneous platforms making it impossible to compute proper estimates. The carbon footprint of each model may be estimated following the protocol defined by T. Parcollet et al. \citep{parcollet21_interspeech}. In practice, it is a function of the PUE of the compute infrastructure, the total energy consumed, and the carbon emission rate of the energy grid. The Jean-Zay supercomputer is energy efficient with a PUE of 0.65 (including heat transfer to nearby infrastructures). We only consider the energy consumed by the GPU and CPU. Power draw measurements are taken every five seconds and then averaged. France's carbon rate is 52 gCO2/kWh \citep{frco2}.

\begin{table}[!th]
\caption{Estimates of the energy in kilowatt hour (kwH) and CO$_2$ equivalent in kilogram produced by the training of the \nameprojectnew models.}
\begin{center}
\begin{tabular}{cc}
        \begin{tabular}{lccccc}
        \toprule
        \textbf{Model} & \textbf{Training time} & \textbf{GPUs} & \textbf{\textcolor{rebuttal}{Total GPU hours}} & \textbf{Energy} & \textbf{CO$_2$} \\  & \textit{(hours)} & & \textit{\textcolor{rebuttal}{(hours)}}  & \textit{(kWh)} & \textit{(kg)} \\
        \hline
        
        1K-\textit{base} & 250 & 4 Tesla V100 & \textcolor{rebuttal}{1,000} & 195.0  & 10.5 \\
        1K-\textit{large} & 925 & 4 Tesla V100 & \textcolor{rebuttal}{3,700} & 721.5 & 37.5   \\
        2.7K-\textit{base} & 128 & 32 Tesla V100 & \textcolor{rebuttal}{4,096} & 682.2 & 35.4  \\
        3K-\textit{base} & 128  & 32 Tesla V100 & \textcolor{rebuttal}{4,096} & 682.2 & 35.4  \\
        3K-\textit{large} & 341  & 32 Tesla V100 & \textcolor{rebuttal}{10,912} & 1,817.5 & 94.5  \\
        7K-\textit{base} & 123  & 64 Tesla V100 & \textcolor{rebuttal}{7,872} & 1,535.0 & 79.8 \\
        7K-\textit{large} & 211  & 64 Tesla V100 & \textcolor{rebuttal}{13,504} & 4,501.0 & 234 \\
        14K-\textit{light} & 156  & 32 Tesla V100 & \textcolor{rebuttal}{4,992} & 1,497.6 & 77.8  \\
        14K-\textit{large} & 436 & 64 Tesla V100 & \textcolor{rebuttal}{27,904} & 8,371.2 & 435  \\
        14K-\textit{xlarge} & 525 & 104 Tesla A100 & \textcolor{rebuttal}{54,600} & 16,511.2 & 859  \\
        \bottomrule\\
    \end{tabular}   
\end{tabular}
\end{center}
\label{tab:summary}
\end{table}

Table \ref{tab:summary} reports the total energy consumed as well as the estimated carbon emissions of all \nameprojectnew models. First, it is quite clear that the carbon footprint of training large SSL models is far from being negligible, even in countries with a relatively clean energy mix. As supported by previous research, the location of the training must be considered as a critical design choice when starting such a project as the total CO$_2$ emissions can easily be multiplied by four to height if gas and oil are integrated in the mix. Finally, one may wonder if the extra kWhs thrown at the model are worth it given the relatively small downstream performance improvement between the 3K-large, 7K-large, and 7K-xlarge, in contrast to the energy consumption being multiplied by a factor 3.5.

\section{Conclusion}
\label{sec:conclusion}

\nameprojectnew establishes new foundations for the development of French SSL-equipped speech technologies. Following the three steps of the lifecycle of any SSL model, we gathered and documented the largest available collection of unlabeled French speech for SSL pre-training, we trained three new pre-trained SSL models for a total of 10 available checkpoints, and we evaluated them on two new tasks increasing the total number of tasks to six. \nameprojectnew models are shared with the community via the HuggingFace Hub.

\section{Acknowledgements}
This work was performed using HPC resources from GENCI–IDRIS (Grants 2022-A0131013821, 2023-A0151014633, AD011013257R2) and 
was also partially supported by MIAI@Grenoble-Alpes (ANR-19-P3IA-0003) and E-SSL project (ANR-22-CE23-0013).
This paper was also partially funded by the European Commission through the SELMA project under grant number 957017, and 
UTTER project under grant number 101070631.
We would like to thank William N. Havard for the original idea of scraping the Audiocite website.

\bibliographystyle{elsarticle-num-names} 
\bibliography{thebib} 


\end{document}